\documentclass[journal]{IEEEtran}
\IEEEoverridecommandlockouts
\usepackage{cite}

\usepackage{amsmath,amssymb,amsfonts,mathtools,bm}
\usepackage{amsthm}
\usepackage{algorithm}
\usepackage{algorithmic}
\usepackage{graphicx}
\usepackage{textcomp}
\usepackage{xcolor,float}
\usepackage{tabularx}
\usepackage{textcomp}
\usepackage{diagbox}
\usepackage{svg}
\usepackage{booktabs}
\usepackage{subfigure}
\usepackage{hyperref}
\usepackage{changes}
\newtheorem{theorem}{Theorem}

\usepackage{booktabs}
\usepackage{array} % for m{...}

\usepackage{multirow}

\usepackage{changes} %宏包

\usepackage{tikz,xcolor,hyperref}% Make Orcid icon
\definecolor{lime}{HTML}{A6CE39}
\DeclareRobustCommand{\orcidicon}{%
    \begin{tikzpicture}
    \draw[lime, fill=lime] (0,0) 
    circle [radius=0.16] 
    node[white] {{\fontfamily{qag}\selectfont \tiny ID}};    \draw[white, fill=white] (-0.0625,0.095) 
    circle [radius=0.007];    \end{tikzpicture}
    \hspace{-2mm}}
\foreach \x in {A, ..., Z}{%
    \expandafter\xdef\csname orcid\x\endcsname{\noexpand\href{https://orcid.org/\csname orcidauthor\x\endcsname}{\noexpand\orcidicon}}
    }
\allowdisplaybreaks
\renewcommand{\baselinestretch}{1}

\begin{document}

% \title{Toward Cross-Receiver Generalization in RF Fingerprint Identification via Feature Disentanglement}
% \title{Receiver-Agnostic RF Fingerprint Identification via Feature Disentanglement and Adversarial Training}
\title{Cross-Receiver Generalization for RF Fingerprint Identification via Feature Disentanglement and Adversarial Training}

% \author{
% \IEEEauthorblockN{
% Yuhao Pan\IEEEauthorrefmark{1},
% Xiucheng Wang\IEEEauthorrefmark{2},
% Wenchao Xu\IEEEauthorrefmark{1},
% Nan Cheng\IEEEauthorrefmark{2}\\
% }
% % \IEEEauthorblockA{
% % \IEEEauthorrefmark{1}Division of Integrative Systems and Design, Hong Kong University of Science and Technology, Hong Kong, China\\
% % \IEEEauthorrefmark{2} School of Telecommunications Engineering, Xidian University, Xi'an, 710071, China\\
% % Email: ypanca@connect.ust.hk, xcwang\_1@stu.xidian.edu.cn, 
% % wenchaoxu@ust.hk, dr.nan.cheng@ieee.org}}
    
%     \maketitle

\author{
Yuhao Pan\orcidA{},
Xiucheng Wang\orcidB{},~\IEEEmembership{Graduate Student Member,~IEEE,}
Fushuo Huo\orcidC{},
Nan Cheng\orcidD{},~\IEEEmembership{Senior Member,~IEEE,}
Wenchao Xu\orcidE{},~\IEEEmembership{Member,~IEEE}
% <-this % stops a space
% 左下角的致谢
\thanks{ }% <-this % stops a space
\thanks{
\par Yuhao Pan and Wenchao Xu are with the Division of Integrative Systems and Design, Hong Kong University of Science and Technology, Hong Kong, China (e-mail: ypanca@connect.ust.hk, wenchaoxu@ust.hk). \textit{Wenchao Xu is the corresponding author}.

\par Xiucheng Wang and Nan Cheng are with the State Key Laboratory of ISN and School of Telecommunications Engineering, Xidian University, Xi’an 710071, China (e-mail: xcwang\_1@stu.xidian.edu.cn, dr.nan.cheng@ieee.org). 

\par Fushuo Huo is with the School of Cyber Science and Engineering, Southeast University, Nanjing, China (e-mail: fushuohuo@seu.edu.cn).

}
}

    \maketitle

% \author{{Yuhao Pan\orcidA{},}
% Xiucheng Wang\orcidB{},~\IEEEmembership{Graduate Student Member,~IEEE,}
% % Jie Gao\orcidC{},~\IEEEmembership{Senior Member,~IEEE,}
% Zhiyao Xu\orcidD{},
% Nan Cheng\orcidE{},~\IEEEmembership{Senior Member,~IEEE,}
% Wenchao Xu\orcidF{},~\IEEEmembership{Member,~IEEE,}
% Jun-jie Zhang\orcidG{}
% % <-this % stops a space
% % 左下角的致谢
% \thanks{ }% <-this % stops a space
% \thanks{
% \par This work was supported by the National Key Research and Development Program of China (2020YFB1807700), and the National Natural Science Foundation of China (NSFC) under Grant No. 62071356. \textit{Nan Cheng is the corresponding author.}
% \par Yuhao Pan is with the School of Electronic Engineering, Xidian University, Xi'an 710071, China (e-mail:yhpan@stu.xidian.edu.cn). \textit{Yuhao Pan and Xiucheng Wang contribute equally.} 
% \par Xiucheng Wang and Nan cheng are with the State Key Laboratory of ISN and School of Telecommunications Engineering, Xidian University, Xi’an 710071, China (e-mail:xcwang\_1@stu.xidian.edu.cn; dr.nan.cheng@ieee.org). 
% \par Zhiyao Xu is with the School of Artificial Intelligence, Xidian University, Xi'an, 710071, China (e-mail:21009200843@stu.xidian.edu.cn).
% \par Wenchao Xu is with the Department of Computing, The Hong Kong Polytechnic University, Hong Kong, China (e-mail: wenchao.xu@polyu.edu.hk).
% \par Jun-jie Zhang is with the Northwest Institute of Nuclear Technology, Xi’an 710024, China (e-mail: zhangjunjie@nint.ac.cn).
% }
% }

%     \maketitle

\IEEEdisplaynontitleabstractindextext

\IEEEpeerreviewmaketitle

\begin{abstract}
Radio frequency fingerprint identification (RFFI) is a key technique for wireless network security, leveraging intrinsic hardware imperfections to enable transmitter identification. Although deep neural networks are effective at extracting discriminative RF features, their performance is significantly affected by receiver-induced variability in practical deployments. 
In real-world scenarios, RF signals inherently entangle transmitter-specific characteristics with receiver-dependent distortions, leading models to capture receiver-related patterns when training and evaluation are conducted on the same device. Consequently, replacing the receiver during deployment often results in notable performance degradation. 
To address this issue, we propose a cross-receiver robust RFFI framework that explicitly disentangles transmitter-specific and receiver-specific representations. The proposed method integrates adversarial domain alignment with receiver-aware regularization to suppress residual receiver information in transmitter features while enforcing intra-receiver consistency in receiver-specific representations. A feature separation constraint is further introduced to decouple the two components in the latent space.
Extensive experiments on multi-receiver WiFi datasets demonstrate that the proposed method consistently outperforms state-of-the-art baselines under cross-receiver evaluation and significantly improves robustness to receiver replacement.

\end{abstract}


\section{Introduction}
In recent years, the rapid advancement of Internet of Things (IoT) technologies has enabled the large-scale deployment of interconnected devices across a wide range of applications~\cite{xu2017internet, lu2014connected}. 
However, this proliferation has also raised growing concerns regarding network security, particularly in wireless environments.
The increasing number of connected devices substantially expands the attack surface of IoT systems (i.e., the set of potential entry points and exploitable vulnerabilities), posing significant challenges for reliably distinguishing legitimate devices from spoofed or unauthorized ones.
Conventional authentication mechanisms, such as Media Access Control (MAC) address verification, are inherently insecure due to their susceptibility to spoofing attacks. 
Meanwhile, cryptography-based security protocols, such as Transport Layer Security (TLS), rely on digital certificates and incur substantial computational overhead, which may exceed the processing capabilities of low-power IoT devices (e.g., smart home sensors), limiting their practicality in resource-constrained scenarios. 
These challenges collectively underscore the need for lightweight, hardware-level authentication mechanisms that can operate effectively in large-scale IoT networks.

One promising solution is Radio Frequency Fingerprint Identification (RFFI), which addresses the aforementioned limitations from a physical-layer perspective. RFFI is founded on intrinsic hardware impairments introduced during the manufacturing of wireless devices. Despite identical design specifications, mass-produced devices exhibit subtle variations in electrical properties, such as resonance frequency, impedance matching, and nonlinear behavior. These physical-layer characteristics serve as unique identifiers, enabling device-level authentication without additional cryptographic overhead. Leveraging this intrinsic property, RFFI has emerged as an effective approach for device identification and classification. In recent years, RFFI has attracted increasing research attention and has also begun to find applications in commercial domains. For instance, in the Automatic Dependent Surveillance Broadcast (ADS-B) system for air traffic control, RFFI has been applied to aircraft identification and classification. This growing interest has catalyzed the technical evolution of RFFI, leading to increasingly sophisticated methods for extracting and exploiting device-specific features.

The development of RFFI techniques can generally be categorized into two main phases: traditional signal processing approaches and deep learning-based methods. 
Early RFFI studies relied on handcrafted waveform- or modulation-domain features extracted from RF signals, which required extensive expert knowledge and limited scalability in end-to-end learning scenarios~\cite{yuan2014specific, ding2018specific, peng2019deep, yang2019multimodal, hua2018accurate}. 
With the advancement of deep learning, RFFI methods based on neural networks—particularly convolutional neural networks (CNNs)—have become mainstream, as they enable direct extraction of discriminative features from raw I/Q signals without manual engineering~\cite{shen2021radio, gopalakrishnan2019robust}. 
Representative efforts include complex-valued CNNs that preserve I/Q inter-channel correlations~\cite{gopalakrishnan2019robust} and the integration of signal compensation techniques such as carrier frequency offset (CFO) correction to enhance feature discriminability~\cite{shen2021radio}. 
Beyond feature learning architectures, several recent studies have explored strategies to improve RFFI robustness under dynamic wireless channels~\cite{al2020exposing, wang2024radiodiff, wang2025radiodiff, sankhe2019oracle, shen2022towards, hanna2022wisig}. 
In parallel, recent studies have explored open-set RFFI to classify known devices and detect unseen ones as unknown. OpenRFI~\cite{yang2025openrfi} serves as a representative method that improves open-set recognition through test-time fine-tuning.
However, existing works predominantly focus on mitigating transmitter--channel variations, while the impact of receiver hardware heterogeneity on RF fingerprint features remains largely underexplored. This limitation poses a critical challenge for real-world cross-receiver deployment.

The receiver-induced bias, though frequently overlooked, represents a critical challenge in practical RFFI systems. Most existing methods are developed and evaluated under the assumption that the same receiver is used for both training and testing. 
This assumption neglects the distribution shift introduced by receiver hardware discrepancies, leading to significant performance degradation when models are deployed on unseen receivers. 
Such limitations pose substantial risks in real-world deployments, where RFFI system functionality may be compromised if the training receiver becomes unavailable due to hardware failures or replacement. 
To address this challenge, we propose a more realistic and demanding cross-receiver generalization scenario: models are trained on data collected from one group of receivers and required to generalize directly to a different set of unseen receivers. This cross-receiver generalization setting is practically important, as it enables seamless model migration and ensures continuous RFFI operation in the event of receiver replacement or malfunction.

Building upon this cross-receiver generalization setting, we propose a robust learning framework termed \emph{Disentangled Representation for Invariant Fingerprint Training (DRIFT)}. 
In our formulation, training data collected from multiple receivers are treated as source domains, while data collected from unseen receivers constitute the target domain, with all domains sharing the same transmitter label space. 
Each source domain corresponds to signals received by a single receiver from multiple transmitters.
The proposed framework first applies channel equalization to suppress channel-induced variations in multi-receiver I/Q signals, and then learns disentangled representations by separating transmitter-specific and receiver-specific feature components. 
Cross-entropy losses are employed to supervise the extraction of both feature types, while a distance-based regularization term is introduced to explicitly enhance their separability in the latent space. 
To promote receiver-invariant transmitter representations, a gradient reversal layer (GRL) is applied to the transmitter-specific features to achieve adversarial alignment across receiver domains. 
For receiver-specific features, we leverage the observation that signals received by the same receiver—regardless of the transmitting device—tend to share consistent receiver characteristics, and accordingly introduce a style-based regularization mechanism to enforce intra-receiver feature consistency.
The main contributions of this paper are summarized as follows:
\begin{enumerate}
    \item We provide a theoretical analysis of cross-receiver RFFI, clarifying how received I/Q signals are jointly influenced by transmitter and receiver hardware characteristics. Based on this analysis, we propose a novel disentangled representation learning framework that explicitly separates transmitter-specific and receiver-specific features.
    
    \item We design a physically motivated learning strategy that incorporates communication-aware priors. Specifically, a GRL is employed to enforce receiver-invariant learning for transmitter-specific features, while a center-based regularization is introduced to encourage intra-receiver consistency of receiver-specific features.
    
    \item We conduct extensive experiments on publicly available multi-receiver RFFI datasets to evaluate the proposed framework. Comprehensive results demonstrate that our method outperforms state-of-the-art approaches and exhibits strong robustness under cross-receiver deployment scenarios.
\end{enumerate}

The remainder of this paper is organized as follows. Section~\ref{sec2} reviews related work. Section~\ref{sec3} formulates the problem and presents the modeling of the optimization objective. Section~\ref{sec4} provides a theoretical analysis to validate the effectiveness of the proposed algorithm. Section~\ref{sec5} introduces the DRIFT architecture. Section~\ref{sec6} evaluates the performance of our method through experiments. Finally, Section~\ref{sec7} concludes the paper and discusses potential directions for future research.

\section{Related Works and Preliminary}
\label{sec2}

Deep learning-based approaches have demonstrated strong performance in RFFI. 
However, in real-world deployments, receiver-specific variations—and to a lesser extent dynamic wireless channels—often induce domain shifts, making it difficult to extract stable and transferable transmitter-specific features. 
This section reviews representative research efforts that aim to improve the robustness and generalization capability of RFFI models under such non-ideal conditions.

\subsection{Deep Learning in RFFI}
Recent research on RFFI has been predominantly driven by deep learning techniques, including CNNs, Long Short-Term Memory (LSTM), attention mechanisms, and Transformers. 
These methods aim to automatically learn discriminative features from raw RF signals for device identification. 
For instance, Das et al.~\cite{das2018deep} employ LSTM networks to model the temporal correlations within I/Q signal streams for the device identification and classification of low-power radio devices. 
Merchant et al. \cite{merchant2018deep} utilize a CNN-based approach on time-domain I/Q signals to achieve high identification and verification accuracy for seven ZigBee devices. 
Peng et al.~\cite{peng2019deep} employ differential constellation trace figures to enable fine-grained device recognition via hardware-level imperfection extraction. 
Furthermore, Zhang et al.~\cite{zhang2022adaptive} propose a dual attention convolutional module that adaptively assigns weights to local and global features of RF fingerprint data by attending to different feature levels. 
To handle varying input lengths of signal samples, Shen et al.~\cite{shen2023toward} adopt a Transformer-based architecture for flexible feature extraction. 
Moreover, Zeng et al.~\cite{zeng2023multi} expand from single-modality to multi-modality input by fusing multiple signal representations (i.e., I/Q samples, carrier frequency offset (CFO), fast Fourier transform (FFT) coefficients, and short-time Fourier transform (STFT) coefficients) via a shared attention mechanism, and concatenate the resulting features to improve classification performance.

\subsection{Domain Adaptation in RFFI}
In machine learning, distribution shifts between training and test datasets often lead to significant performance degradation of models trained solely on source domains when evaluated on unseen target domains. To address this issue, domain adaptation (DA) and domain generalization (DG) techniques have been extensively studied to enhance model robustness across diverse domains. Building upon these advances, researchers in wireless communications have employed DA methods to alleviate the performance deterioration in RFFI tasks caused by time-varying wireless channels and receiver-dependent variations.
\subsubsection{Domain Adaptation for Channel Variability}
Pan et al.~\cite{pan2024equalization} adopted correlation alignment (CORAL) to compensate for residual channel effects that remain after channel equalization. Chai et al.~\cite{chai2024channel} proposed a multi-task learning framework with multiple classifiers to improve model robustness under varying channel conditions. Chen et al.~\cite{chen2024prototype} introduced a Prototype-based Domain Discrepancy Alignment (PDDA) method to mitigate distribution shifts in RFFI. 
Zhao et al.~\cite{zhao2024cross} further advanced domain adaptation by integrating a prototype model with few-shot learning to enhance generalization to unseen domains. 
Wang et al.~\cite{wang2024feature} proposed FATransformer to address cross-domain RFFI from an unsupervised domain adaptation perspective by aligning intermediate Transformer features between source and target domains to reduce distribution shifts.
Moreover, Wan et al.~\cite{wan2024vc} proposed a robust emitter identification framework employing adversarial training and semantic consistency to extract channel-invariant features through semi-supervised domain adaptation.

\subsubsection{Domain Adaptation for Cross-Receiver RFFI}
In addition to channel variability, cross-receiver RFFI introduces additional challenges due to receiver-specific variations that induce cross-domain distribution shifts. To address this, several studies have investigated cross-receiver domain adaptation techniques. Zha et al.~\cite{zha2023cross} leveraged the SimSiam framework for unsupervised pretraining and incorporated local maximum mean discrepancy (LMMD) as a regularization term for feature alignment to improve generalization across receivers. Chen et al.~\cite{chen2024cross} proposed a domain adaptation-based method for cross-receiver RFFI, employing a receiver discriminator to perform supervised domain alignment based on receiver labels. These methods typically assume access to unlabeled data from the target receiver domain, which enables domain adaptation during deployment.

\subsection{Domain Generalization in RFFI}
DA leverages source domain data, source domain labels, and target domain data during training to align the distributions of source and target domains. In contrast, DG relies solely on source domain data and labels without requiring access to any target domain information during training, making it more practical yet inherently more challenging. Wang et al.~\cite{wang2025avoiding} proposed a single-source DG approach that employs random overlay augmentation (ROA) for domain expansion and dual alignment via contrastive learning, enabling extraction of transmitter features invariant to channel variations. Recent studies have further explored DG for cross-receiver RFFI scenarios. Shen et al.~\cite{shen2023towards} introduced a receiver classifier branch and employed a GRL to enforce the robustness of transmitter feature extraction to variations across different receivers. 
% Zhao et al.~\cite{zhao2023gan} proposed an efficient cross-receiver identification framework by pretraining the transmitter feature extraction module and subsequently fine-tuning the module with data from new receivers. 
Zhao et al.~\cite{zhao2023gan} proposed a receiver-agnostic transmitter fingerprinting framework that learns separate receiver- and transmitter-related representations using dual branches and a distance-based regularization, and deploys the learned transmitter feature extractor across unseen receivers.
Zhang et al.~\cite{zhang2024domain} developed a model that disentangles extracted features into transmitter-independent and receiver-independent components, trained using a combination of cross-entropy loss (CE), information entropy loss (IE), and mutual independence loss (MI).
Zhou et al.~\cite{zhou2025receiver} proposed a receiver-agnostic RFFI baseline that separates mixed features into orthogonal components via a binary mask separator, and enhances receiver-invariant transmitter feature learning through cross-combination and an auxiliary reconstruction constraint.

Compared with existing cross-receiver DG methods for RFFI, our DRIFT explicitly disentangles transmitter- and receiver-related representations using an MSE-based separation regularizer. In particular, we enforce receiver invariance on transmitter-related features via a GRL, while regularizing receiver-specific features with a center-based loss, resulting in robust transmitter representations for identification.

\begin{figure*}[ht]
    \centering
    \includegraphics[width=1.80\columnwidth]{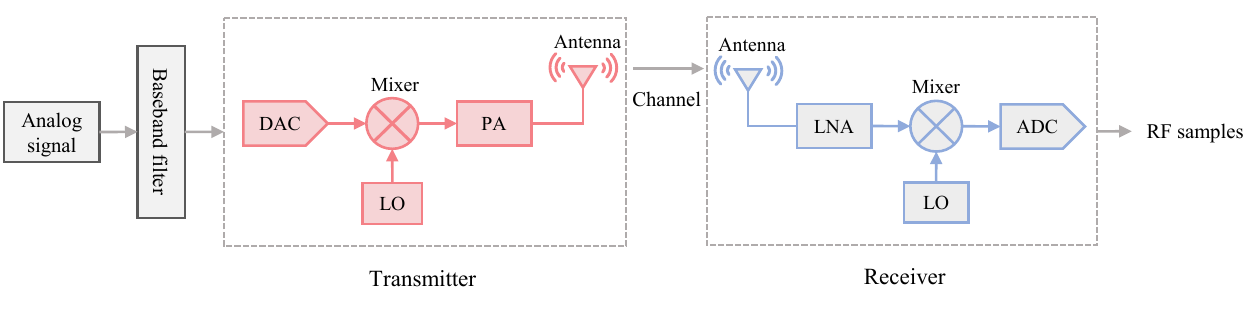}
    \caption{Illustration of transmitter and receiver hardware impairments.}
    \label{hardware impairments}
\end{figure*}

\section{System Model and Problem Formulation}
\label{sec3}

\subsection{RF Fingerprint Modeling Across Receivers}
In this subsection, we present an RF fingerprint modeling process across different receivers and analyze the key factors influencing the RFFI task, as illustrated in Fig.~\ref{hardware impairments}.

During transmission, the analog signal is first sampled by an analog-to-digital converter (ADC) to produce a discrete digital signal that is subsequently modulated into a digital baseband signal. This digital signal is then converted into an analog baseband signal by a digital-to-analog converter (DAC), upconverted into an RF signal via a local oscillator (LO), and finally amplified by a power amplifier (PA) before transmission. During signal propagation, channel effects can further distort the transmitted signal. On the receiver side, the incoming RF signal is first amplified by a low-noise amplifier (LNA), downconverted by an LO, and digitized again by an ADC. The resulting I/Q samples serve as the input for RFFI. Therefore, in a cross-receiver RFFI task, the I/Q signal $X_{ij}^t$ received by receiver $j$ from transmitter $i$ at time $t$ can be modeled as:

\begin{equation}
X_{ij}^t = g_j\!\Big( h_{ij}^t\!\big( f_i(s_i(t)) \big) \Big)
\label{obj1}
\end{equation}
where $s_i(t)$ denotes the baseband signal generated by transmitter $i$ at time $t$; 
$f_i$ represents the hardware impairments of transmitter $i$, such as DAC quantization noise and power amplifier (PA) nonlinear distortion; 
$h_{ij}^t$ denotes the time-varying channel response between transmitter $i$ and receiver $j$, including multipath fading and ambient noise; 
$g_j$ denotes the receiver-specific hardware impairments of receiver $j$, such as ADC quantization noise and LNA nonlinear distortion.

It is worth noting that the formulation in \eqref{obj1} does not aim to explicitly model each individual RF hardware impairment at the circuit or sample level.
Instead, we adopt an \emph{effective baseband representation} that captures the \emph{aggregate influence} of transmitter and receiver hardware characteristics on the received I/Q signal.
In practical RFFI scenarios, transmitter- and receiver-side hardware impairments jointly introduce device-dependent distortions that exhibit relatively stable statistical patterns under fixed hardware configurations.
From the perspective of fingerprint learning, these combined effects can be reasonably approximated as multiplicative factors acting on the original baseband signal, which are absorbed into the transmitter-specific term $f_i$ and the receiver-specific term $g_j$ in \eqref{obj1}.
This abstraction allows us to focus on learning invariant and disentangled representations for cross-receiver RFFI, rather than performing fine-grained physical modeling of individual RF impairments.

\subsection{Problem Formulation}
In cross-receiver RFFI, both transmitters and receivers exhibit unique hardware-specific fingerprint characteristics, as illustrated in \eqref{obj1}. Receiver-specific features induce significant domain shifts between different receiver domains, since the same transmitter's signal may exhibit distinct distributions when captured by different receivers. Specifically, for the $m$-th receiver, the collected dataset $\mathcal{D}_s^m$ inherently contains receiver-specific hardware impairment features $\mathbf{g}_m$ (consistent with \eqref{obj1}), which interfere with reliable transmitter identification.

During training, we assume that I/Q signal samples are collected from $M$ edge receiver nodes in an IoT network, with each sample originating from one of $K$ transmitters. The entire training dataset is defined as
\begin{equation}
\mathcal{D}_s = \{\mathcal{D}_s^1, \mathcal{D}_s^2, \ldots, \mathcal{D}_s^M\},
\end{equation}
where $\mathcal{D}_s^m$ denotes the sub-dataset collected by the $m$-th receiver, given by
\begin{equation}
\mathcal{D}_s^m = \left\{ (x_{m,i}, y_{m,i}, d_{m,i}) \right\}_{i=1}^{N_m},
\end{equation}
where $N_m$ is the total number of samples collected by the $m$-th receiver, $x_{m,i}$ denotes the $i$-th I/Q signal sample from the $m$-th receiver, $y_{m,i} \in \{1, \ldots, K\}$ is the corresponding transmitter label, and $d_{m,i} = m$ specifies the receiver domain label (i.e., the index of the receiver that collects the sample). 
Each sample is thus represented as a labeled tuple comprising the signal data, the transmitter identity label, and the receiver domain label. These receiver-specific sub-datasets are collectively utilized to train a model that generalizes to previously unseen receiver domains.

To enhance generalization performance in cross-receiver RFFI, our objective is to learn a robust model that is insensitive to receiver-specific hardware characteristics $\mathbf{g}_m$. 
This implies that features learned from data collected by $M$ known receivers should still enable reliable transmitter identification when deployed on an unseen $(M+1)$-th receiver domain. 
Accordingly, the model is expected to effectively extract transmitter-specific fingerprint features while suppressing the influence of receiver-specific characteristics.
Under this formulation, we consider a neural network parameterized by $\theta$, which consists of a feature extractor followed by a transmitter classifier $f(\cdot)$. 
The classifier is designed to capture intrinsic transmitter-specific representations while being robust to receiver-induced variations. 
For generality, the training objective can be expressed as the minimization of the following loss function:
\begin{equation}
\theta^\ast = \arg\min_{\theta} 
\mathcal{L}_{\mathrm{CE}}\!\left(f(x; \theta), y\right) 
+ \sum_{j} \lambda_j \mathcal{L}_{\mathrm{reg}}^{(j)}(\theta),
\end{equation}
where $\theta$ denotes the model parameters, $\mathcal{L}_{\mathrm{CE}}$ is the standard cross-entropy loss for transmitter classification, $\mathcal{L}_{\mathrm{reg}}^{(j)}$ represents the $j$-th regularization objective for enhancing cross-receiver generalization, and $\lambda_j$ controls the relative contribution of each regularization term.

\section{Domain Generalization Theory-Based Analysis}
\label{sec4}
Traditional supervised learning methods hinge on the independent and identically distributed (i.i.d.) assumption, where training and testing data follow the same underlying distribution. In contrast, domain generalization (DG) targets models that generalize to unseen target domains using multiple labeled source domains, with the core premise that source and target domains share the same label space but exhibit non-negligible distribution discrepancies. The central challenge of DG is to learn robust models that maintain performance on distribution-shifted target domains, even when no target domain data is accessible during training.
To provide a theoretical foundation for the proposed DRIFT framework, we first recap the fundamental concepts of DG, then establish a generalization risk upper bound under our problem setting, and finally connect the derived bound to the design of DRIFT.

\subsection{Theoretical Analysis}
\subsubsection{Notation}
Let $\{ \mathcal{D}_S^1, \mathcal{D}_S^2, \ldots, \mathcal{D}_S^n \}$ denote multiple source domains, $X$ the input space, $Y$ the label space, and $f: X \rightarrow Y$ the ground-truth labeling function. 
We define a hypothesis space $\mathcal{H}$ consisting of prediction functions $h: X \rightarrow Y$, and the risk of $h$ under distribution $\mathcal{D}$ as
\begin{equation}
R[h] = \mathbb{E}_{x \sim \mathcal{D}} \left[ L(h(x), f(x)) \right],
\end{equation}
where $L: Y \times Y \rightarrow \mathbb{R}_+$ is a non-negative loss function.

\subsubsection{$\mathcal{H}$-Divergence}
The $\mathcal{H}$-divergence quantifies the distributional discrepancy between two domains~\cite{kifer2004detecting}. 
For the multi-class RFFI task, we adopt the standard $\mathcal{H}$-divergence as a domain discrepancy metric, where binary indicator functions induced by the multi-class hypothesis space are used to distinguish samples from different domains.
For hypothesis space $\mathcal{H}$, the $\mathcal{H}$-divergence between $\mathcal{D}_1$ and $\mathcal{D}_2$ is defined as
\begin{equation}
d_{\mathcal{H}}(\mathcal{D}_1, \mathcal{D}_2) = 2 \sup_{h \in \mathcal{H}} 
\left| 
\Pr_{x \sim \mathcal{D}_1}[ h(x) = 1 ] - 
\Pr_{x \sim \mathcal{D}_2}[ h(x) = 1 ] 
\right|.
\end{equation}

\subsubsection{Source Domain Convex Hull}
For multiple source domains, we define the mixture of source distributions as the convex hull of the source domains, denoted by $\Lambda_S$. This convex hull consists of all weighted combinations of source domain distributions:
\begin{equation}
\Lambda_S = 
\left\{
\bar{\mathcal{D}}: \bar{\mathcal{D}}(\cdot) = \sum_{i=1}^{n} \pi_i \mathcal{D}_S^i(\cdot)
\;\middle|\;
\pi = (\pi_1,\ldots,\pi_n) \in \Delta_{n-1}
\right\},
\end{equation}
where $\Delta_{n-1}$ is the $n-1$-dimensional simplex ($\pi_i \geq 0$ and $\sum_{i=1}^{n} \pi_i = 1$). 
For an unseen target domain $\mathcal{D}_U$, we find the optimal $\bar{\mathcal{D}}_U \in \Lambda_S$ that minimizes $d_{\mathcal{H}}(\mathcal{D}_U, \bar{\mathcal{D}}_U)$:
\begin{equation}
\bar{\mathcal{D}}_U = 
\arg \min_{\pi_1,\ldots,\pi_n} 
d_{\mathcal{H}} \left( \mathcal{D}_U, \sum_{i=1}^{n} \pi_i \mathcal{D}_S^i \right),
\end{equation}
which is the closest convex combination of sources to the target.
With the above definitions, we next present the generalization upper bound for the risk on an unseen target domain.

\begin{theorem}[Generalization Bound on Unseen Domain Risk]
\label{theorem-1}
Under the above setting, let $S$ be the set of source domains.
For an unseen target domain distribution $\mathcal{D}_U$, the risk of a hypothesis $h \in \mathcal{H}$ on $\mathcal{D}_U$, denoted as $R_U[h]$, is upper bounded as follows \cite{albuquerque2019generalizing, zhao2019learning, ben2010theory}:
\begin{equation}
\begin{aligned}
R_U[h] &\leq \sum_{i=1}^{n} \pi_i R_S^i[h] + \gamma + \epsilon  \\
&\quad + \min \left\{ 
\mathbb{E}_{\bar{\mathcal{D}}_U} \left[ |f_{S_\pi} - f_U| \right], 
\mathbb{E}_{\mathcal{D}_U} \left[ |f_U - f_{S_\pi}| \right] 
\right\},
\end{aligned}
\end{equation}
\end{theorem}
Here, $R_S^i[h]$ denotes the risk of $h$ on the $i$-th source domain $\mathcal{D}_S^i$. The function $f_{S_\pi}(x) = \sum_{i=1}^{n} \pi_i f_{S_i}(x)$ represents the weighted combination of label functions from all source domains. The term $\gamma = d_{\mathcal{H}}(\bar{\mathcal{D}}_U, \mathcal{D}_U)$ is the $\mathcal{H}$-divergence between the target domain $\mathcal{D}_U$ and its closest convex combination $\bar{\mathcal{D}}_U$ within the convex hull of source domains. 
The term $\epsilon$ denotes the maximum $\widetilde{\mathcal{H}}$-divergence among all domains in the source domain set $S$, where $\widetilde{\mathcal{H}} = \left\{ \mathrm{sign} \left( \left| h(x) - h^\prime(x) \right| - t \right) \mid h,h^\prime \in \mathcal{H},\ 0 \leq t \leq 1 \right\}$.
Intuitively, $\widetilde{\mathcal{H}}$ captures the disagreement patterns between pairs of hypotheses in $\mathcal{H}$, and the resulting $\widetilde{\mathcal{H}}$-divergence measures the worst-case discrepancy among source domains.
Finally, $\mathbb{E}_{\bar{\mathcal{D}}_U}[|f_{S_\pi} - f_U|]$ and $\mathbb{E}_{\mathcal{D}_U}[|f_U - f_{S_\pi}|]$ measure the deviation between the unseen domain label function $f_U$ and the aggregated source label function $f_{S_\pi}$.

\subsection{Effectiveness of DRIFT}
Based on the established generalization bound, we further simplify Theorem~\ref{theorem-1} under the shared-labeling-function condition and then connect the simplified bound to the design of DRIFT. Specifically, assume:
\begin{equation}
f_{S_1} = f_{S_2} = \cdots = f_{S_n} = f_U.
\end{equation}
Here, the shared labeling function refers to the invariant mapping from transmitter-specific RF characteristics to transmitter identities, rather than identical observed I/Q distributions across domains.
Cross-day temporal channel variations mainly perturb the marginal distribution of received signals, while the underlying transmitter identity of each sample remains unchanged.
When this shared-label condition is mildly violated in practical deployments, the label-discrepancy term in Theorem~\ref{theorem-1} becomes nonzero, leading to a looser but still meaningful upper bound.
Under this assumption, and given that \( \sum_{i=1}^{n} \pi_i = 1 \), the aggregated label function \( f_{S_\pi}(x) \), defined as a weighted combination of source domain label functions, simplifies to
\begin{equation}
f_{S_\pi}(x) = \sum_{i=1}^{n} \pi_i f_{S_i}(x) = \sum_{i=1}^{n} \pi_i f_U(x) = f_U(x).
\end{equation}
This implies that
\begin{equation}
|f_{S_\pi}(x) - f_U(x)| = 0.
\end{equation}
Consequently, the label discrepancy term in Theorem \ref{theorem-1} becomes zero:
\begin{equation}
\min\left\{ 
\mathbb{E}_{\bar{\mathcal{D}}_U}\left[ \left|f_{S_\pi}(x) - f_U(x)\right| \right], 
\ \mathbb{E}_{\mathcal{D}_U}\left[ \left|f_U(x) - f_{S_\pi}(x)\right| \right] 
\right\} = 0.
\end{equation}
Therefore, the risk upper bound on the unseen target domain reduces to
\begin{equation}
R_U[h] \le \sum_{i=1}^{n} \pi_i R_S^i[h] + \gamma + \epsilon.
\label{obj}
\end{equation}

This simplified bound provides a theoretical foundation for DRIFT-based domain generalization, indicating that when source and target domains share the same labeling function, the generalization ability on unseen domains can be improved by reducing $\gamma$ and $\epsilon$. 
This theoretical insight directly motivates the core design of DRIFT, i.e., a feature disentanglement strategy to extract transmitter-specific features while suppressing domain-specific interference introduced by receivers. Following this strategy, the disentangled feature spaces for transmitters in the source and target domains are denoted by $\{ \widetilde{\mathcal{D}}_S^1, \widetilde{\mathcal{D}}_S^2, \ldots, \widetilde{\mathcal{D}}_S^n \}$ and $\widetilde{\mathcal{D}}_U$, respectively.
Through this separation, the source-domain transmitter-related features are encouraged to become more domain-invariant, which helps reduce the source-domain discrepancy term $\epsilon$. 
Moreover, by suppressing receiver-specific components, the unseen target-domain feature distribution is expected to become closer to the convex hull of source-domain feature distributions, thereby helping reduce $\gamma$.
These effects provide a principled explanation for the improved generalization performance achieved by our disentanglement-based approach.
The detailed design of the proposed framework is presented in Section~\ref{sec5}.

% 为了排版，把这张图提前了
\begin{figure*}[ht]
  \centering
  \includegraphics[width=1.65\columnwidth]{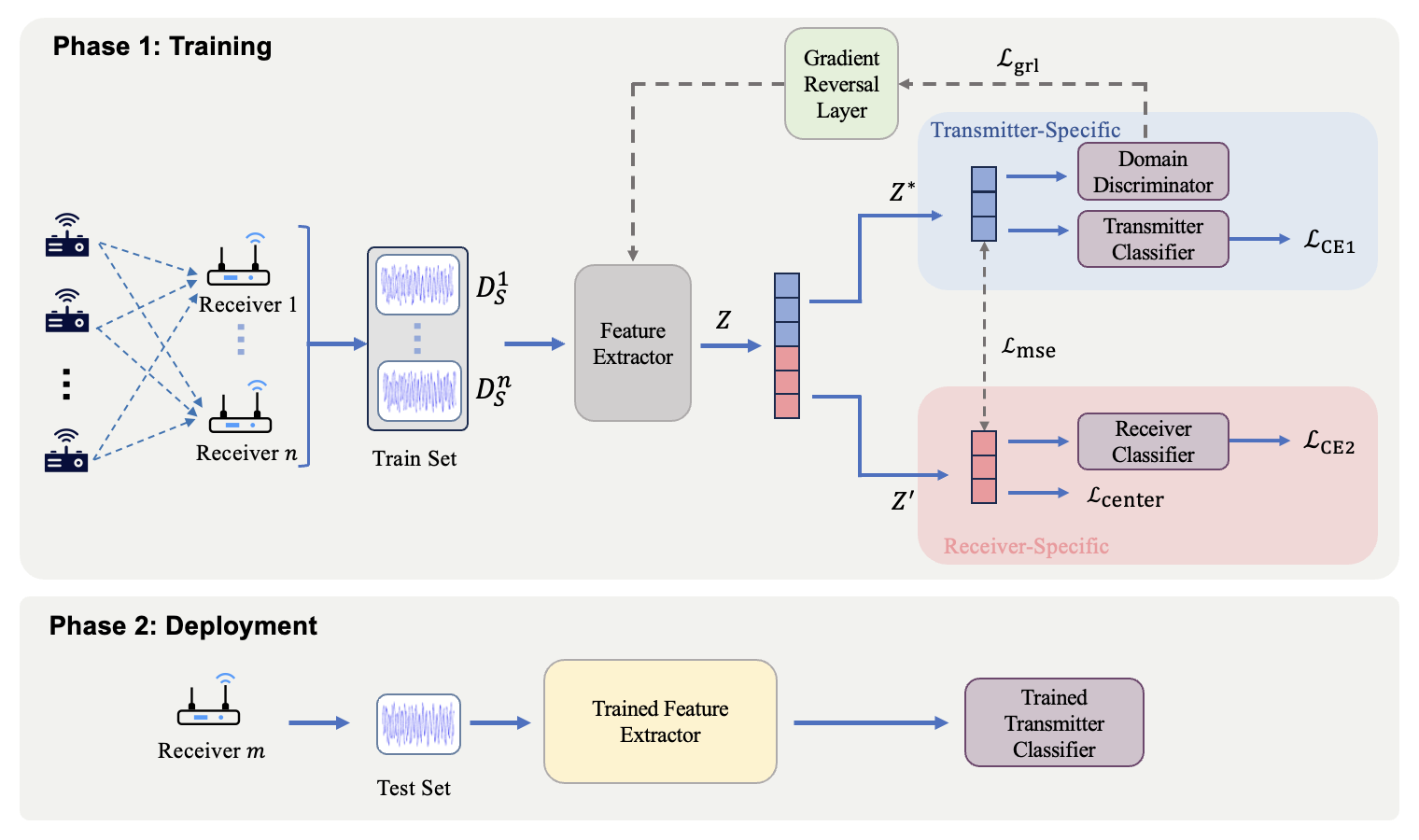}
  \centering \caption{Overview of the proposed DRIFT framework for cross-receiver RFFI, illustrating the training and deployment phases. During deployment, only the trained feature extractor and transmitter classifier are used for inference on unseen receivers.} 
  \label{system model}
   % \vspace{-9pt}
\end{figure*}

\section{Model Architecture and Design}
\label{sec5}
Fig.~\ref{system model} presents the overall architecture of DRIFT. 
The received signal $\mathbf{x}$ is first passed through the feature extraction module $W$, yielding a latent representation $\mathbf{z} = W(\mathbf{x})$. 
The representation $\mathbf{z}$ is then disentangled into two components, corresponding to transmitter-specific and receiver-specific features, denoted as $\mathbf{z}^\ast$ and $\mathbf{z}^\prime$, respectively.
To achieve the objective defined in~\eqref{obj}, we employ cross-entropy loss functions to supervise the extraction of both $\mathbf{z}^\ast$ and $\mathbf{z}^\prime$. 
In addition, a mean squared error (MSE) loss is introduced to explicitly increase the distance between $\mathbf{z}^\ast$ and $\mathbf{z}^\prime$ in the feature space, thereby enhancing their separability. 
To further align transmitter features $\mathbf{z}^\ast$ across different source domains and suppress residual receiver-specific information, we incorporate adversarial training using a GRL during the extraction of $\mathbf{z}^\ast$.
Moreover, to enhance the extraction of receiver-specific features $\mathbf{z}^\prime$ and boost the disentanglement effect, we impose an additional regularization term $\mathcal{L}_{\text{center}}$ on $\mathbf{z}^\prime$.

\subsection{Feature Extraction Module}
In recent years, various deep neural network backbones have been applied to RFFI, including AlexNet, VGG, and ResNet \cite{he2016deep}. Among these, ResNet effectively alleviates the degradation problem in deep neural networks through residual connections, which enable stable training of deeper models and enhance feature extraction capabilities. 
For this reason, we adopt ResNet-18 as the backbone for feature extraction in our framework. Considering that the collected I/Q signals are sequential and structured as $2 \times 256$ samples, we replace the 2D convolutional layers in ResNet-18 with 1D convolutional layers to effectively capture fine-grained temporal dependencies. 
Specifically, the original Conv2D and BatchNorm2D layers are replaced with Conv1D and BatchNorm1D layers to adapt the network to one-dimensional inputs. The architecture first applies a $7 \times 1$ convolution followed by max-pooling to reduce the signal length and capture local temporal structures. 
Subsequently, four groups of residual blocks extract hierarchical representations. 
Finally, a global average pooling layer generates a fixed-length embedding vector $\mathbf{z} \in \mathbb{R}^d$, where $d$ denotes the dimension of the embedding space, and $\mathbf{z}$ represents the deep semantic features of the input signal.

To disentangle transmitter-specific and receiver-specific features, we adopt a simple dimension-wise feature partitioning mechanism.
The extracted $d$-dimensional feature vector $\mathbf{z}$ is partitioned into two equal sub-vectors: the first $d/2$ dimensions correspond to transmitter-specific features $\mathbf{z}^\ast$, and the remaining $d/2$ dimensions correspond to receiver-specific features $\mathbf{z}^\prime$. 
The feature extraction process is formally defined as:
\begin{equation}
    \mathbf{z}^\ast, \mathbf{z}^\prime = f_{\mathrm{split}}\left( f_{\mathrm{emd}}(\mathbf{x}; \theta) \right),
\end{equation}
where $\mathbf{x}$ denotes the input I/Q signal, $ \theta $ represents the learnable parameters of the feature extractor, $ f_{\mathrm{emd}}(\cdot; \theta) $ denotes the ResNet-18-based 1D feature extraction function, and $ f_{\mathrm{split}}(\cdot) $ denotes the dimension-wise feature partitioning operation.

\subsection{Transmitter and Receiver Classification Losses}
To guide the learning of disentangled feature representations, cross-entropy losses are applied to the transmitter and receiver classifiers.
Specifically, the transmitter-specific feature $\mathbf{z}^\ast$ and receiver-specific feature $\mathbf{z}^\prime$, extracted from the feature extraction module, are fed into a transmitter classifier $f(\cdot)$ and a receiver classifier $h(\cdot)$, respectively.
Each classifier comprises three fully connected (FC) layers, and a softmax activation is applied to the output logits to obtain the corresponding class probability distributions.
The predicted labels for transmitters and receivers are then computed as
\begin{align}
    \hat{y} &= \arg\max_j f(\mathbf{z}^\ast)_j, \\
    \hat{d} &= \arg\max_j h(\mathbf{z}^\prime)_j,
\end{align}
where $\hat{y}$ and $\hat{d}$ represent the predicted transmitter and receiver labels, respectively.

The overall classification loss is defined as the sum of transmitter and receiver classification losses, calculated via standard cross-entropy:
\begin{equation}
    \mathcal{L}_{\text{CE}} =
    \mathcal{L}_{\text{CE1}}\big(f(\mathbf{z}^\ast), y\big)
    +
    \mathcal{L}_{\text{CE2}}\big(h(\mathbf{z}^\prime), d\big),
\end{equation}
where $y$ and $d$ denote the ground-truth transmitter and receiver labels.
This objective provides supervised signals for learning transmitter- and receiver-discriminative representations, which facilitates the subsequent disentanglement regularization and improves generalization.

\subsection{Transmitter-Specific Feature Regularization}
Although we employ an explicit feature separation strategy to extract transmitter-specific features $\mathbf{z}^\ast$, residual receiver-related information may still be retained in $\mathbf{z}^\ast$, which degrades the domain invariance of transmitter features across different receiver domains. To further optimize $\mathbf{z}^\ast$ and eliminate such residual receiver-specific information, we integrate a Gradient Reversal Layer (GRL) \cite{ganin2016domain} into our framework to align $\mathbf{z}^\ast$ across receiver domains.

The GRL is tailored here to the optimization of $\mathbf{z}^\ast$: it acts as an identity mapping during the forward pass (i.e., $\mathrm{GRL}(\mathbf{z}^\ast) = \mathbf{z}^\ast$) and does not alter the feature flow, but reverses the gradient direction during backpropagation by multiplying the gradient by a negative coefficient $-\lambda$ (a fixed or tunable scalar). 
This gradient reversal mechanism enables adversarial training, in which the receiver domain discriminator is encouraged to correctly identify receiver domains in the forward pass, while the feature extractor is adversarially optimized to suppress receiver-discriminative information in the backward pass, thereby enforcing receiver domain invariance in $\mathbf{z}^\ast$.
Specifically, we insert the GRL between $\mathbf{z}^\ast$ and a receiver domain discriminator $D$ (implemented as a two-layer fully connected network), where $D$ outputs a probability distribution over receiver domains to distinguish the origin domain of $\mathbf{z}^\ast$. The adversarial loss for optimizing $\mathbf{z}^\ast$ is defined as:
\begin{equation}
    \mathcal{L}_{\mathrm{grl}} 
    = \mathbb{E}_{(\mathbf{z}^\ast,\, d)} 
    \left[ - \log D\!\left(\mathrm{GRL}(\mathbf{z}^\ast)\right)_{d} \right],
\end{equation}
where $d \in \{1, \dots, |\mathcal{D}|\}$ denotes the receiver domain label of the input sample, and $\mathcal{D}$ is the set of all training receiver domains. Formally, the GRL operation for $\mathbf{z}^\ast$ is defined as:
\begin{align}
    \mathrm{GRL}(\mathbf{z}^\ast) &= \mathbf{z}^\ast, \quad \text{(forward pass)}, \\
    \frac{\partial \mathrm{GRL}(\mathbf{z}^\ast)}{\partial \mathbf{z}^\ast} &= -\lambda I, \quad \text{(backward pass)},
\end{align}
where $\lambda$ is a coefficient that controls the strength of gradient reversal. This adversarial training with GRL suppresses residual receiver-specific information in $\mathbf{z}^\ast$, thereby enhancing the receiver invariance of $\mathbf{z}^\ast$ while preserving its transmitter-discriminative characteristics.

\subsection{Receiver-Specific Feature Regularization}
Given the unique characteristics of I/Q signals in communication scenarios, distinct differences exist between cross-receiver RFFI and traditional image domain generalization. Traditional domain generalization primarily addresses variations in visual styles (e.g., sketches, cartoons, real-world images). However, in the cross-receiver RFFI task within communication systems, after applying channel equalization to mitigate the impact of channel noise, the primary differences in data distribution arise from the hardware impairments of different receivers.

In this context, I/Q signals from multiple transmitters received by the same receiver can be categorized into the same receiver domain. 
Since receiver and transmitter hardware impairments affect the I/Q baseband signal in similar ways, the influence of receiver-specific attributes on signals from different transmitters tends to exhibit consistent patterns, resembling a style transfer effect. 
Consequently, the receiver-specific features $\mathbf{z}^\prime$ extracted from samples of different transmitters within the same receiver domain should exhibit high similarity. 
Based on this observation, we introduce an auxiliary regularization task with a center-based constraint to regularize the receiver-specific features $\mathbf{z}^\prime$, encouraging features from the same receiver domain to cluster around a shared centroid.
The corresponding center-based loss is defined as:
\begin{equation}
    \mathcal{L}_{\text{center}} = \sum_{d \in \mathcal{D}} \frac{1}{\left|S_d\right|} \sum_{i \in S_d} \left\| \mathbf{z}_i^\prime - \mathbf{c}_d \right\|_2^2
\end{equation}
where $\mathcal{D}$ denotes the set of all receiver domains, $S_d$ is the index set of samples belonging to receiver domain $d$, $\mathbf{z}_i^\prime$ is the receiver-specific feature of the $i$-th sample in domain $d$, and $\mathbf{c}_d = \frac{1}{|S_d|} \sum_{j \in S_d} \mathbf{z}_j^\prime$ denotes the feature centroid of receiver domain $d$. This $\mathcal{L}_{\text{center}}$ regularization enhances intra-domain compactness of $\mathbf{z}^\prime$, thereby complementing the disentanglement of transmitter- and receiver-specific features.

\subsection{Feature Separation Loss}
To explicitly encourage transmitter-specific and receiver-specific features to diverge in the latent space, we introduce a feature separation loss $\mathcal{L}_{\text{mse}}$ based on negative squared Euclidean distance, where minimizing $\mathcal{L}_{\text{mse}}$ is equivalent to maximizing the distance between the two feature components.
Formally, the loss is defined as
\begin{equation}
    \mathcal{L}_{\text{mse}} = - \frac{1}{N} \sum_{i=1}^{N} 
    \left\| \mathbf{z}^\ast_i - \mathbf{z}^\prime_i \right\|_2^2,
\end{equation}
where $N$ denotes the number of samples in a mini-batch, $\mathbf{z}^\ast_i$ and $\mathbf{z}^\prime_i$ represent the transmitter-specific and receiver-specific features of the $i$-th sample, respectively.

This constraint penalizes feature similarity between $\mathbf{z}^\ast$ and $\mathbf{z}^\prime$, encouraging them to encode complementary and non-overlapping information. 
When combined with classification losses, adversarial regularization, and receiver-specific feature constraints, $\mathcal{L}_{\text{mse}}$ reinforces effective feature disentanglement and improves the model’s cross-receiver generalization performance.

\subsection{Optimization via Backpropagation}
Based on the previously introduced loss components, including the cross-entropy classification loss $\mathcal{L}_{\mathrm{CE}}$, the GRL-based adversarial loss $\mathcal{L}_{\mathrm{grl}}$, the receiver-specific feature regularization loss $\mathcal{L}_{\mathrm{center}}$, and the feature separation loss $\mathcal{L}_{\mathrm{mse}}$, the overall objective function of the proposed model is formulated as:
\begin{equation}
    \mathcal{L} = \mathcal{L}_{\mathrm{CE}} + \lambda_1 \mathcal{L}_{\mathrm{grl}} + \lambda_2 \mathcal{L}_{\mathrm{center}} + \lambda_3 \mathcal{L}_{\mathrm{mse}},
\end{equation}
where $\lambda_1$, $\lambda_2$, and $\lambda_3$ are hyperparameters that control the contributions of the corresponding loss terms to the total objective.
During training, gradients of $\mathcal{L}$ with respect to the feature extractor parameters $\theta$ are computed via standard backpropagation, and all network parameters are jointly optimized using gradient-based optimization.
Algorithm~\ref{alg:drift_training} summarizes the overall training procedure of DRIFT.

\begin{algorithm}[htbp]
\caption{Training Procedure of DRIFT}
\label{alg:drift_training}
\begin{algorithmic}[1]
\REQUIRE Training dataset $\mathcal{D}_s = \{ \mathcal{D}_s^1, \ldots, \mathcal{D}_s^M \}$ ($M$: number of receiver domains); 
model parameters $\theta$; learning rate $\eta$; total epochs $T$; batch size $B$; 
hyperparameters $\lambda_1, \lambda_2, \lambda_3$
\ENSURE Trained model parameters $\theta^\ast$

\FOR{epoch $= 1$ to $T$}
    \FOR{each mini-batch $\{(x_i, y_i, d_i)\}_{i=1}^B$ sampled from $\mathcal{D}_s$}

        \STATE \textit{Feature extraction and splitting:}
        \STATE \hspace{5mm} $\mathbf{z}_i \gets f_{\mathrm{emd}}(x_i; \theta)$
        \STATE \hspace{5mm} $\mathbf{z}_i^\ast, \mathbf{z}_i^\prime \gets f_{\mathrm{split}}(\mathbf{z}_i)$

        \STATE \textit{Classification losses:}
        \STATE \hspace{5mm} $\mathcal{L}_{\mathrm{CE1}} \gets \mathrm{CE}(f(\mathbf{z}^\ast), y)$
        \STATE \hspace{5mm} $\mathcal{L}_{\mathrm{CE2}} \gets \mathrm{CE}(h(\mathbf{z}^\prime), d)$
        \STATE \hspace{5mm} $\mathcal{L}_{\mathrm{CE}} \gets \mathcal{L}_{\mathrm{CE1}} + \mathcal{L}_{\mathrm{CE2}}$

        \STATE \textit{Adversarial loss via GRL:}
        \STATE \hspace{5mm} $\mathcal{L}_{\mathrm{grl}} \gets \mathrm{CE}\big(D(\mathrm{GRL}(\mathbf{z}^\ast)), d\big)$

        \STATE \textit{Receiver-specific center regularization:}
        \STATE \hspace{5mm} For each receiver domain $d$ appearing in the mini-batch:
        \STATE \hspace{10mm} $S_d \gets \{ i \mid d_i = d \}$, \quad $\mathbf{c}_d \gets \frac{1}{|S_d|}\sum_{j\in S_d}\mathbf{z}_j^\prime$
        \STATE \hspace{5mm} $\mathcal{L}_{\mathrm{center}} \gets \sum_{d} \frac{1}{|S_d|} \sum_{i \in S_d} \left\| \mathbf{z}_i^\prime - \mathbf{c}_d \right\|_2^2$

        \STATE \textit{Feature separation loss (negative MSE):}
        \STATE \hspace{5mm} $\mathcal{L}_{\mathrm{mse}} \gets -\frac{1}{B}\sum_{i=1}^B \left\| \mathbf{z}_i^\ast - \mathbf{z}_i^\prime \right\|_2^2$

        \STATE \textit{Total loss and update:}
        \STATE \hspace{5mm} $\mathcal{L} \gets \mathcal{L}_{\mathrm{CE}} + \lambda_1 \mathcal{L}_{\mathrm{grl}} + \lambda_2 \mathcal{L}_{\mathrm{center}} + \lambda_3 \mathcal{L}_{\mathrm{mse}}$
        \STATE \hspace{5mm} Update $\theta$ via backpropagation and gradient-based optimization

    \ENDFOR
\ENDFOR
\RETURN $\theta^\ast$
\end{algorithmic}
\end{algorithm}

\section{Experiment}
\label{sec6}
\subsection{Experimental Setup}
This subsection summarizes the dataset, experimental protocol, baseline methods, and implementation details used in our evaluation.

\subsubsection{Datasets}
We utilize the open-source WiSig dataset~\cite{hanna2022wisig}, a large-scale publicly available dataset widely used for radio frequency fingerprinting research. The dataset contains approximately 10 million I/Q signal packets collected from 174 WiFi transmitters and 41 USRP receivers at four different time points over the course of one month. To facilitate systematic evaluation, the dataset is organized into four subsets: \emph{ManySig}, \emph{ManyTx}, \emph{ManyRx}, and \emph{SingleDay}. Data collection was conducted on four evenly spaced days within the same month, with 1{,}000 I/Q signal samples recorded per transmitter--receiver (Tx--Rx) pair on each day. All transmitters are based on the Atheros AR5212/AR5213 chipset, while the receivers include devices from the USRP N210 and USRP B210 series. 
Notably, both transmitters and receivers are labeled according to their spatial coordinates $(x,y)$ on the ORBIT testbed, a $20 \times 20$ grid with 3-foot (approximately 1-meter) spacing between nodes. 
For example, a receiver located at coordinate $(1,1)$ is denoted as $Rx_{(1,1)}$. 
In this work, transmitter signals observed by different receivers are treated as distinct domains, which naturally facilitates the study of cross-receiver domain generalization in RFFI.

\subsubsection{Experimental Protocol}
The experiments are conducted on the ManySig dataset, which is collected in a controlled indoor testbed environment. 
Transmitters and receivers are deployed within a confined indoor space, resulting in relatively stable propagation conditions compared to outdoor or mobile scenarios. 
Under this setting, temporal channel variations are moderate, allowing us to focus on receiver-induced domain shifts in RF fingerprint identification.

We apply channel equalization as the only signal-level preprocessing step, following prior RFFI studies~\cite{hanna2022wisig}. 
Signal synchronization, frequency offset correction, and MMSE-based channel equalization are performed using the MATLAB WLAN Toolbox (R2019b) with default parameters. 
Signals are downsampled to 20~Msps for processing and resampled to 25~Msps after equalization. 
The corrected frequency offset is then reapplied to preserve fingerprint-related characteristics.

\begin{table*}[t]
    \centering
    \caption{Performance (\%) of All Methods on Different Receiver Combinations}
    \label{tab:performance}
    \renewcommand{\arraystretch}{1.12}
    \setlength{\tabcolsep}{7pt}
    \small
    \begin{tabular}{@{}
        >{\centering\arraybackslash}m{0.46\textwidth}
        c c c c c
    @{}}
        \toprule
        \textbf{Training Receivers} & \textbf{ERM} & \textbf{DANN} & \textbf{RIEI} & \textbf{Zhou} & \textbf{DRIFT} \\
        \midrule

        $Rx_{(1,1)},\, Rx_{(14,7)}$
        & 59.54 & 63.25 & 60.64 & \underline{\textcolor[rgb]{0.0,0.2,0.7}{63.66}}
        & \textbf{\textcolor[rgb]{0.7,0.0,0.0}{68.66}} \\

        $Rx_{(1,1)},\, Rx_{(1,19)}$
        & 64.61 & 62.89 & 68.90 & \textbf{\textcolor[rgb]{0.7,0.0,0.0}{71.91}}
        & \underline{\textcolor[rgb]{0.0,0.2,0.7}{69.73}} \\

        $Rx_{(1,1)},\, Rx_{(14,7)},\, Rx_{(7,7)}$
        & \underline{\textcolor[rgb]{0.0,0.2,0.7}{68.13}}
        & 58.70 & 63.66 & 67.58 & \textbf{\textcolor[rgb]{0.7,0.0,0.0}{73.54}} \\

        $Rx_{(1,1)},\, Rx_{(1,19)},\, Rx_{(7,7)}$
        & \underline{\textcolor[rgb]{0.0,0.2,0.7}{72.42}}
        & 63.83 & 68.21 & 69.76 & \textbf{\textcolor[rgb]{0.7,0.0,0.0}{75.49}} \\

        $Rx_{(1,1)},\, Rx_{(14,7)},\, Rx_{(18,2)},\, Rx_{(7,7)}$
        & \underline{\textcolor[rgb]{0.0,0.2,0.7}{70.38}}
        & 62.60 & 65.73 & 70.37 & \textbf{\textcolor[rgb]{0.7,0.0,0.0}{71.28}} \\

        $Rx_{(1,1)},\, Rx_{(1,19)},\, Rx_{(14,7)},\, Rx_{(8,8)}$
        & 69.57 & 65.32 & 66.55 & \underline{\textcolor[rgb]{0.0,0.2,0.7}{72.33}}
        & \textbf{\textcolor[rgb]{0.7,0.0,0.0}{76.62}} \\

        $Rx_{(1,1)},\, Rx_{(1,19)},\, Rx_{(14,7)},\, Rx_{(7,7)},\, Rx_{(8,8)}$
        & 77.20 & 67.56 & 64.50 & \underline{\textcolor[rgb]{0.0,0.2,0.7}{80.64}}
        & \textbf{\textcolor[rgb]{0.7,0.0,0.0}{80.82}} \\

        $Rx_{(1,19)},\, Rx_{(14,7)},\, Rx_{(19,2)},\, Rx_{(20,1)},\, Rx_{(8,8)}$
        & \underline{\textcolor[rgb]{0.0,0.2,0.7}{72.33}}
        & 70.72 & 69.50 & 72.18 & \textbf{\textcolor[rgb]{0.7,0.0,0.0}{77.61}} \\

        \bottomrule
    \end{tabular}
\end{table*}

\subsubsection{Baselines}
To evaluate the effectiveness of the proposed algorithm, we compare it against four representative baseline methods.

\textbf{Baseline 1: ERM.}
Empirical Risk Minimization (ERM) optimizes model parameters by minimizing the average classification loss over the training data and serves as a non-domain-generalization baseline.

\textbf{Baseline 2: DANN.}
Proposed by Ganin et al.~\cite{ganin2016domain}, this method adopts an adversarial learning strategy from the general domain adaptation literature to enhance domain generalization.

\textbf{Baseline 3: RIEI.}
Proposed by Zhang et al.~\cite{zhang2024domain}, RIEI disentangles the extracted features into transmitter-related and receiver-related representations, which are fed into separate classifiers. MI loss and IE loss are introduced to facilitate effective feature disentanglement. During inference, only the transmitter-related features and the corresponding classifier are used.

\textbf{Baseline 4: Zhou et al.}
This baseline~\cite{zhou2025receiver} achieves receiver-agnostic RFFI by first extracting mixed transmitter-receiver features, then segregating them into orthogonal components via a binary mask separator. Disentangled features are cross-combined, with a reconstruction module to preserve transmitter information, and optimized end-to-end.

All baseline methods are trained and evaluated under the same experimental protocol to ensure a fair comparison.

\subsubsection{Implementation Details}
We train the proposed method using the Adam optimizer with a batch size of 256 and an initial learning rate of 0.0001. The hyperparameters of DRIFT are set as $\lambda_1 = 1$, $\lambda_2 = 0.01$, and $\lambda_3 = 0.02$, unless otherwise specified.
All experiments are implemented in PyTorch 2.0.0 with Python 3.8, and conducted on a system equipped with an Intel Xeon Platinum 8255C CPU and an NVIDIA GeForce RTX 2080Ti GPU. 
To ensure fairness and reproducibility, we conduct experiments with 5 distinct random seeds. For each run corresponding to a specific seed, we save the model checkpoint from the final epoch. We then report the average performance across the 5 independent runs to mitigate the impact of randomness.

\subsection{Experiment 1: Cross-Receiver Generalization}
To validate the effectiveness of the proposed DRIFT algorithm for cross-receiver domain generalization, we conduct comprehensive comparisons on the ManySig dataset against four representative baselines, including ERM, DANN, RIEI, and the method proposed by Zhou \emph{et al.} 
To isolate receiver-induced domain shifts from temporal channel variations, all experiments in this subsection are restricted to data collected on Day~1.
For training, we randomly sample 800 I/Q samples per receiver; when $n$ receivers are used, the total number of training samples is $6 \times 800 \times n$. During evaluation, models are tested independently on all remaining unseen receivers (not used in training), where we randomly sample 200 I/Q samples per receiver for testing, and the final performance is reported as the average accuracy across these receivers.

We begin with scenarios involving two training receivers. As shown in Table~\ref{tab:performance}, DRIFT achieves the best performance of 68.66\% for the training set $\{Rx_{(1,1)}, Rx_{(14,7)}\}$, which is 5.01\% higher than the second-best method (Zhou \emph{et al.}); for $\{Rx_{(1,1)}, Rx_{(1,19)}\}$, DRIFT obtains the second-best performance of 69.73\%, only 2.18\% lower than the optimal method and still competitive with all baselines. These results indicate that DRIFT can effectively generalize even under limited training receiver diversity.

We then increase the training receiver diversity to three and four receivers. In these settings, DRIFT consistently achieves the highest average accuracy across unseen receivers, with 73.54\% for the three-receiver set $\{Rx_{(1,1)}, Rx_{(14,7)}, Rx_{(7,7)}\}$ and a new high of 76.62\% for the four-receiver set $\{Rx_{(1,1)}, Rx_{(1,19)}, Rx_{(14,7)}, Rx_{(8,8)}\}$. Moreover, the performance gap between DRIFT and the baseline methods becomes more pronounced as additional receivers are incorporated into training (e.g., 5.41\% higher than ERM for the three-receiver set and 4.29\% higher than Zhou \emph{et al.} for the above four-receiver set), suggesting that DRIFT is able to more effectively exploit multi-receiver information to learn receiver-invariant and transmitter-discriminative representations.

Further experiments under more diverse four-receiver and five-receiver training configurations confirm this trend. 
For the four-receiver set $\{Rx_{(1,1)}, Rx_{(14,7)}, Rx_{(18,2)}, Rx_{(7,7)}\}$, DRIFT achieves the highest accuracy of 71.28\%. 
Under five-receiver training scenarios, DRIFT consistently attains the best performance, reaching 80.82\% for $\{Rx_{(1,1)}, Rx_{(1,19)}, Rx_{(14,7)}, Rx_{(7,7)}, Rx_{(8,8)}\}$ and 77.61\% for $\{Rx_{(1,19)}, Rx_{(14,7)}, Rx_{(19,2)}, Rx_{(20,1)}, Rx_{(8,8)}\}$. 
These results further demonstrate the robustness and scalability of DRIFT as training domain diversity increases.

\begin{table*}[t]
    \centering
    \caption{Cross-Day Performance (\%) of All Methods on Different Receiver Combinations}
    \label{tab:cross_day_performance}
    \renewcommand{\arraystretch}{1.2}
    \setlength{\tabcolsep}{8pt}
    \small
    \begin{tabular}{c c c c c c c}
        \toprule
        \textbf{Training Receivers} &
        \textbf{Test Receiver (Day 1 $\rightarrow$ Day n)} &
        \textbf{ERM} & \textbf{DANN} & \textbf{RIEI} & \textbf{Zhou} & \textbf{DRIFT} \\
        \midrule

        % ------------------ 2 receivers ------------------
        \multirow{3}{*}{$Rx_{(1,1)},\, Rx_{(14,7)}$}
        & Day 2 &
        60.44 & 64.12 &
        \underline{\textcolor[rgb]{0.0,0.2,0.7}{64.99}} & 64.32 &
        \textbf{\textcolor[rgb]{0.7,0.0,0.0}{68.15}} \\
        & Day 3 &
        62.19 & 62.84 &
        \underline{\textcolor[rgb]{0.0,0.2,0.7}{65.00}} & 62.66 &
        \textbf{\textcolor[rgb]{0.7,0.0,0.0}{70.28}} \\
        & Day 4 &
        60.10 & 60.44 &
        \underline{\textcolor[rgb]{0.0,0.2,0.7}{63.43}} & 61.80 &
        \textbf{\textcolor[rgb]{0.7,0.0,0.0}{64.90}} \\
        \midrule

        % ------------------ 3 receivers ------------------
        \multirow{3}{*}{$Rx_{(1,1)},\, Rx_{(14,7)},\, Rx_{(7,7)}$}
        & Day 2 &
        71.68 & 67.28 & 69.63 &
        \underline{\textcolor[rgb]{0.0,0.2,0.7}{75.37}} &
        \textbf{\textcolor[rgb]{0.7,0.0,0.0}{77.21}} \\
        & Day 3 &
        72.47 & 69.96 & 72.11 &
        \underline{\textcolor[rgb]{0.0,0.2,0.7}{72.54}} &
        \textbf{\textcolor[rgb]{0.7,0.0,0.0}{78.33}} \\
        & Day 4 &
        71.04 & 69.33 & 70.91 &
        \underline{\textcolor[rgb]{0.0,0.2,0.7}{71.23}} &
        \textbf{\textcolor[rgb]{0.7,0.0,0.0}{79.48}} \\
        \midrule

        % ------------------ 4 receivers ------------------
        \multirow{3}{*}{$Rx_{(1,1)},\, Rx_{(14,7)},\, Rx_{(18,2)},\, Rx_{(7,7)}$}
        & Day 2 &
        65.55 & 64.47 & 65.92 &
        \underline{\textcolor[rgb]{0.0,0.2,0.7}{66.25}} &
        \textbf{\textcolor[rgb]{0.7,0.0,0.0}{73.82}} \\
        & Day 3 &
        64.89 & 64.05 & 63.62 &
        \underline{\textcolor[rgb]{0.0,0.2,0.7}{65.58}} &
        \textbf{\textcolor[rgb]{0.7,0.0,0.0}{70.85}} \\
        & Day 4 &
        63.41 & 61.01 & 65.67 &
        \textbf{\textcolor[rgb]{0.7,0.0,0.0}{67.24}} &
        \underline{\textcolor[rgb]{0.0,0.2,0.7}{65.97}} \\
        \midrule

        % ------------------ 5 receivers ------------------
        \multirow{3}{*}{$Rx_{(1,1)},\, Rx_{(1,19)},\, Rx_{(14,7)},\, Rx_{(7,7)},\, Rx_{(8,8)}$}
        & Day 2 &
        70.64 & 67.94 & 66.97 &
        \underline{\textcolor[rgb]{0.0,0.2,0.7}{75.47}} &
        \textbf{\textcolor[rgb]{0.7,0.0,0.0}{77.90}} \\
        & Day 3 &
        71.83 & 71.23 & 69.77 &
        \underline{\textcolor[rgb]{0.0,0.2,0.7}{72.79}} &
        \textbf{\textcolor[rgb]{0.7,0.0,0.0}{82.22}} \\
        & Day 4 &
        70.04 & 64.82 & 65.06 &
        \underline{\textcolor[rgb]{0.0,0.2,0.7}{71.03}} &
        \textbf{\textcolor[rgb]{0.7,0.0,0.0}{76.02}} \\

        \bottomrule
    \end{tabular}
\end{table*}

Overall, DRIFT achieves the best performance in 7 out of 8 training receiver combinations and ranks second in the remaining case, with a performance gap of less than 2.2\%. These results indicate that DRIFT maintains consistent and competitive cross-receiver generalization under single-day conditions, outperforming existing baseline methods across a wide range of training configurations.

\textbf{Visualization Analysis.}
We further analyze the cross-receiver generalization behavior from two complementary perspectives: confusion matrices for classification-level errors and t-SNE visualization for feature-space alignment.

\begin{figure}[ht]
  \centering
  \subfigure[ERM]{
    \includegraphics[width=0.46\columnwidth]{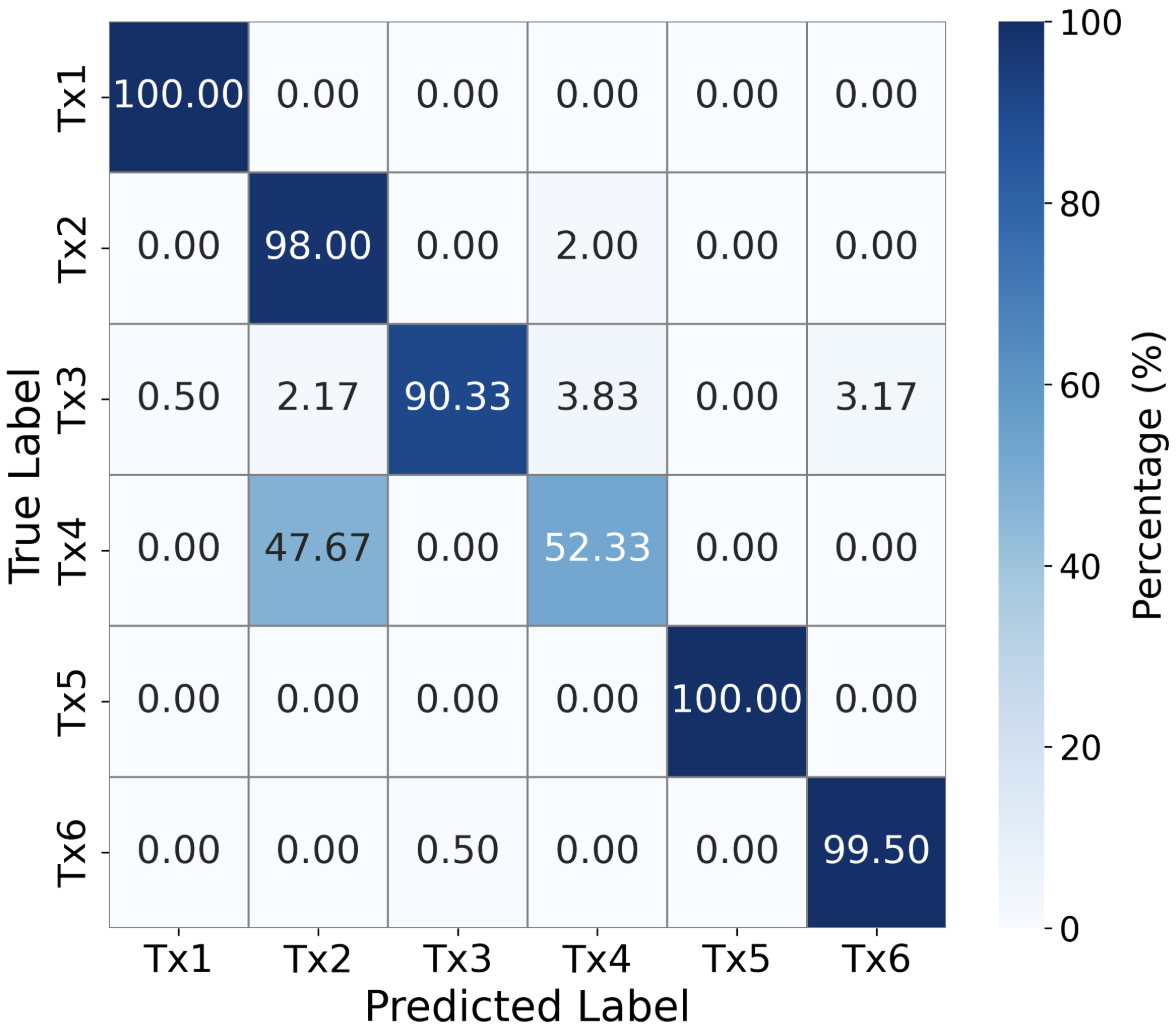}
  }
  \subfigure[DRIFT]{
    \includegraphics[width=0.46\columnwidth]{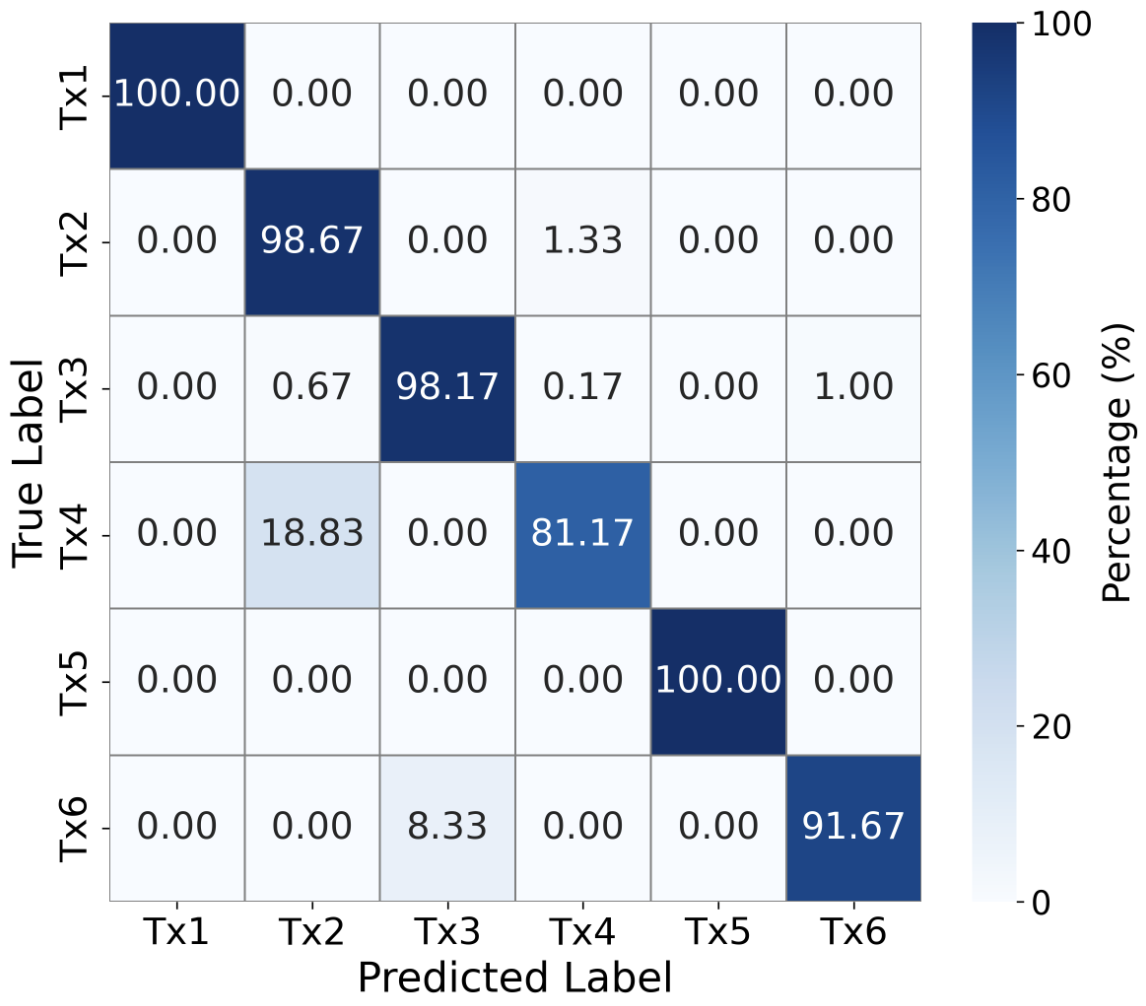}
  }
  \caption{Confusion matrices of ERM and DRIFT under the cross-receiver setting.}
  \label{fig:confusion_comparison}
\end{figure}

First, we visualize the confusion matrices of ERM and DRIFT in Fig.~\ref{fig:confusion_comparison}.
The models are trained on five receivers $\{Rx_{(1,1)}, Rx_{(1,19)}, Rx_{(14,7)}, Rx_{(7,7)}, Rx_{(8,8)}\}$ and evaluated on the unseen receiver $Rx_{(7,14)}$.
ERM shows noticeable off-diagonal confusions among several transmitter classes, indicating sensitivity to receiver-induced domain shifts.
In contrast, DRIFT yields a more concentrated diagonal structure with substantially fewer misclassifications.
These observations are consistent with the quantitative results and further validate the effectiveness of the proposed feature disentanglement and adversarial domain alignment for cross-receiver RFFI.

\begin{figure}[ht]
  \centering
  \subfigure[ERM]{
    \includegraphics[width=0.43\columnwidth]{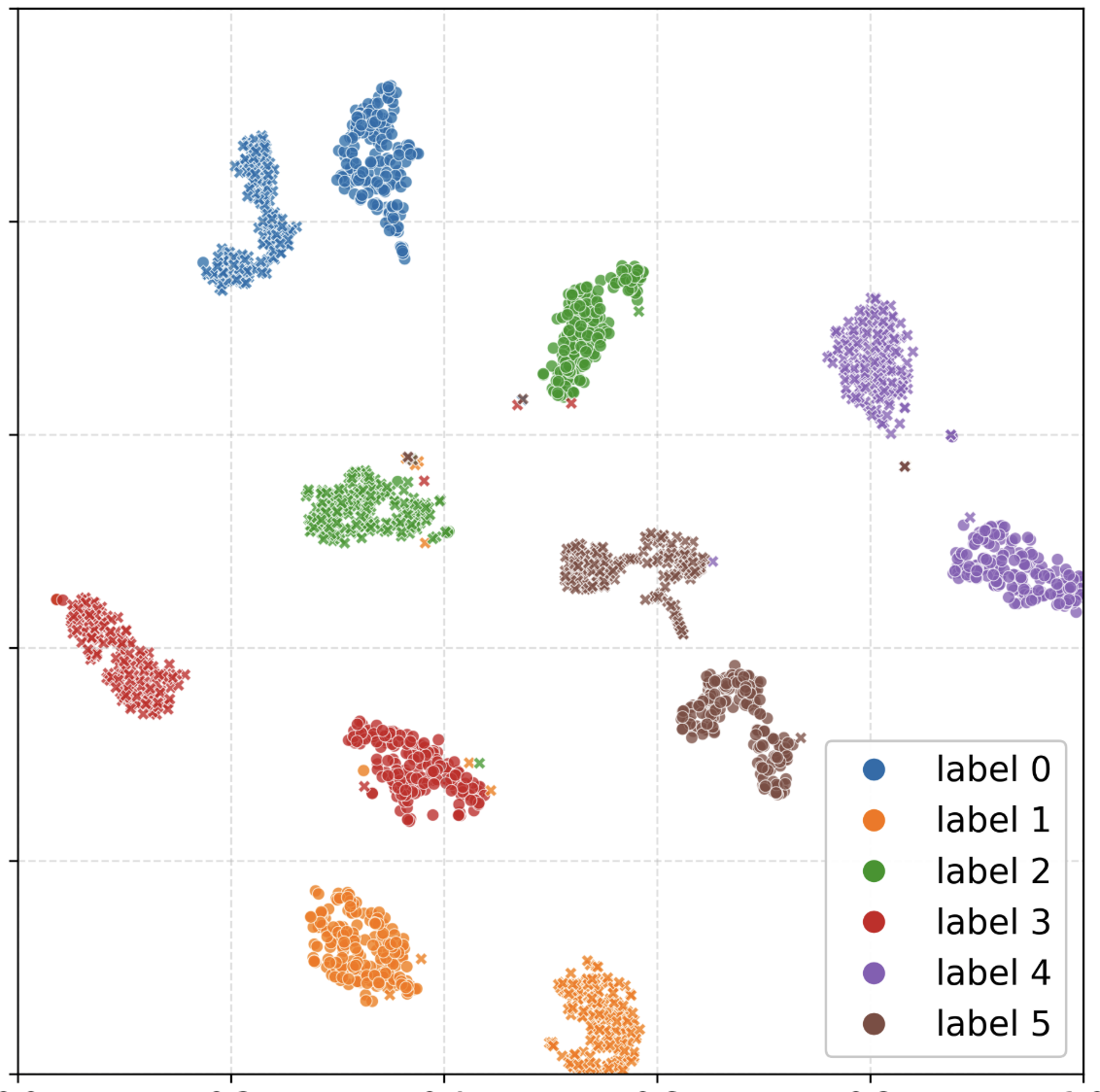}
  }
  \subfigure[DRIFT]{
    \includegraphics[width=0.43\columnwidth]{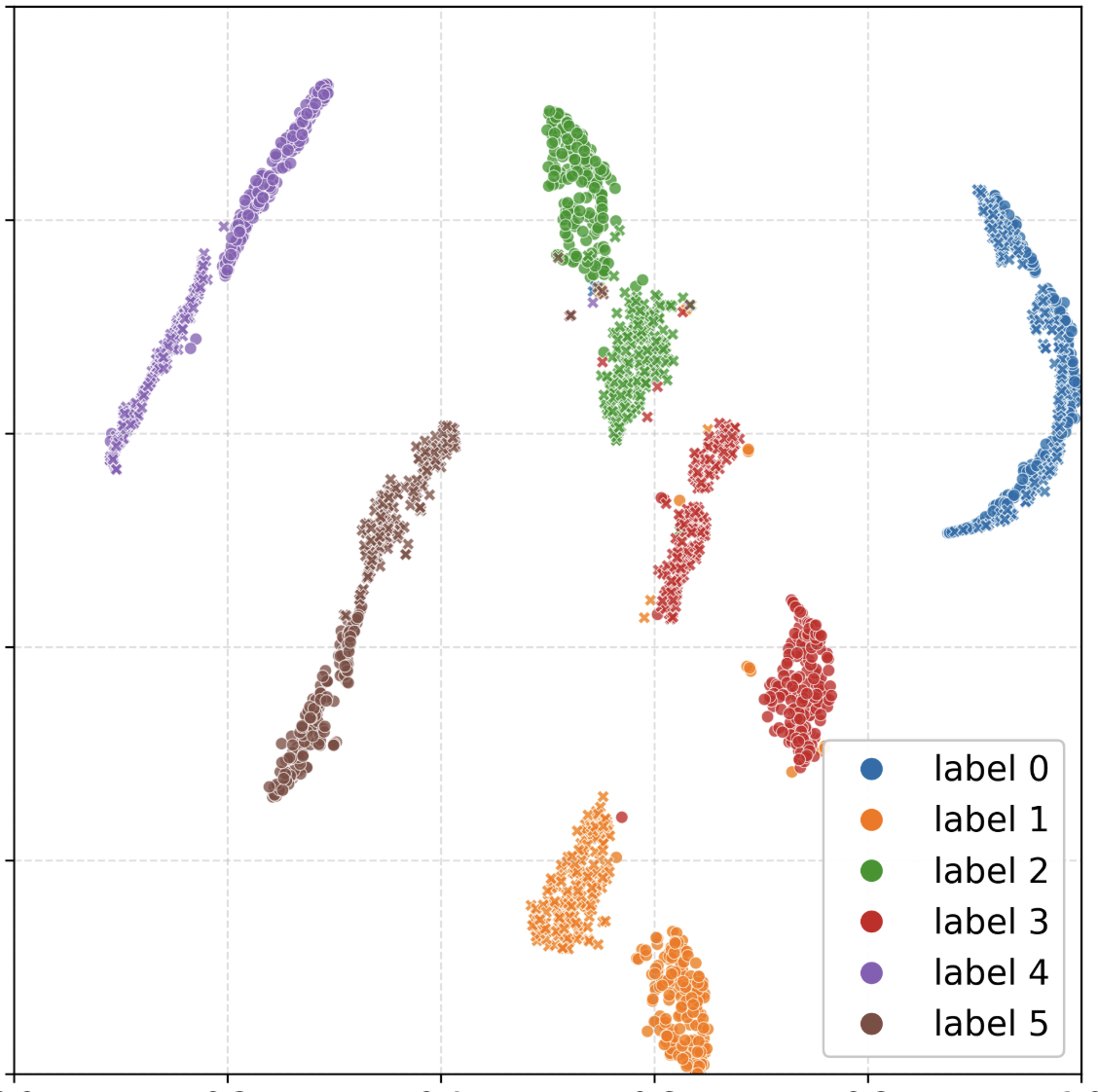}
  }
  \caption{t-SNE visualization of transmitter-related features $\mathbf{z}^\ast$ learned by ERM and DRIFT under the cross-receiver setting. Circles/crosses denote samples from two unseen test receivers; colors denote transmitter classes.}
  \label{fig:Tsne}
\end{figure}

Second, we further visualize the transmitter-related features $\mathbf{z}^\ast$ learned by DRIFT and ERM using t-SNE in Fig.~\ref{fig:Tsne}.
The models are trained on $\{Rx_{(1,1)},\, Rx_{(1,19)},\, Rx_{(14,7)},\, Rx_{(7,7)},\, Rx_{(8,8)}\}$ and evaluated on the unseen receivers $Rx_{(7,14)}$ (circles) and $Rx_{(19,2)}$ (crosses), with colors indicating transmitter classes.
Compared with ERM, DRIFT yields more compact intra-class clusters and better cross-receiver alignment: features of the same transmitter from different receivers overlap more closely, while different classes remain more separable.

\subsection{Experiment 2: Cross-Day Model Stability}
In this set of experiments, we further evaluate the cross-receiver generalization performance of different methods under data collected on different days. Models are trained using signals from Day~1 and evaluated on data collected on Day~2, Day~3, and Day~4, while keeping the training and testing receivers strictly disjoint. For each target day, performance is reported as the average transmitter identification accuracy over all unseen receivers.

We first consider the two-receiver training configuration $\{Rx_{(1,1)}, Rx_{(14,7)}\}$. As shown in Table~\ref{tab:cross_day_performance}, DRIFT consistently achieves the highest accuracy across all test days, reaching 68.15\%, 70.28\%, and 64.90\% on Day~2, Day~3, and Day~4, respectively. Compared with the strongest competing baseline (RIEI), DRIFT maintains a clear and stable performance advantage of 1.47\% to 5.28\% across all three days despite the limited training receiver diversity.
When the number of training receivers is increased to three and four, the overall performance of all methods improves, indicating that increased receiver diversity during training benefits cross-day generalization. In the three-receiver configuration, DRIFT achieves an average accuracy of 78.34\%, outperforming the second-best Zhou baseline by a concrete margin of 5.29\%. For the four-receiver configuration, DRIFT remains superior on Day~2 (7.57\% higher than Zhou) and Day~3 (5.27\% higher than Zhou), while the Zhou baseline slightly outperforms DRIFT on Day~4 by a narrow margin of 1.27\%, suggesting that disentanglement-based methods may benefit from specific receiver combinations under certain conditions but exhibit less consistent behavior across days.
In the most challenging setting with five training receivers, DRIFT again achieves the highest accuracy on all test days, peaking at 82.22\% on Day~3 (surpassing the Zhou baseline by 9.43\%) and maintaining a consistent advantage over all baselines, with an average accuracy of 78.71\% across the three days. These results indicate that DRIFT effectively leverages increased training domain diversity to enhance robustness under cross-day evaluation.

Overall, the cross-day evaluation demonstrates that DRIFT provides consistently strong and stable performance, achieving the highest average accuracy across all experimental configurations. This confirms that DRIFT is capable of learning transmitter-discriminative and receiver-invariant representations.

\subsection{Experiment 3: Contribution of Each Module}
In this section, we conduct an ablation study to analyze the contribution of each key component in the proposed DRIFT framework to cross-receiver generalization. Training is performed using the receiver set $\{Rx_{(1,1)},\, Rx_{(1,19)},\, Rx_{(14,7)},\, Rx_{(7,7)},\, Rx_{(8,8)}\}$, and evaluation is conducted on five unseen receivers $\{Rx_{(19,2)},\, Rx_{(2,1)},\, Rx_{(2,19)},\, Rx_{(20,1)},\, Rx_{(7,14)}\}$ across four different days (Day~1 to Day~4). 
The \textit{Basic Model} includes only transmitter and receiver classifiers trained using standard cross-entropy loss, without explicit feature separation or regularization, achieving an average accuracy of 75.09\%. We then progressively introduce additional components, including the gradient reversal layer (GRL), center regularization (Cen), and the proposed MSE-based separation module, as well as their combinations.

\begin{table}[t]
    \centering
    \caption{Ablation Study of Each Component in the proposed DRIFT}
    \label{tab:ablation_drift}
    \renewcommand{\arraystretch}{1.2}
    \setlength{\tabcolsep}{5pt}
    \begin{tabular}{lccccc}
        \toprule
        \textbf{Model} & \textbf{Day 1} & \textbf{Day 2} & \textbf{Day 3} & \textbf{Day 4} & \textbf{Avg} \\
        \midrule
        Basic Model 
        & 77.39 & 72.69 & 76.01 & 74.28 & 75.09 \\
        \midrule
        +GRL 
        & 72.33 & 72.54 & 74.67 & 70.55 & 72.53 \\
        +Cen 
        & 76.08 & 73.77 & 76.96 & 73.76 & 75.14 \\
        +MSE 
        & 77.35 & 74.25 & 80.43 & 75.68 & 76.93 \\
        +MSE+GRL 
        & 79.02 & 76.71 & 78.73 & 73.79 & 77.56 \\
        +MSE+Cen 
        & 76.79 & 73.35 & 80.29 & 74.95 & 76.85 \\
        +GRL+Cen 
        & 72.46 & 76.25 & 79.58 & 73.84 & 75.53 \\
        \midrule
        \textbf{Full Model} 
        & \textbf{80.82} & \textbf{77.90} & \textbf{82.22} & \textbf{76.02} & \textbf{79.24} \\
        \bottomrule
    \end{tabular}
\end{table}

As shown in Table~\ref{tab:ablation_drift}, the effects of different modules vary across settings. While introducing GRL alone leads to a performance drop to 72.53\%, both center regularization and the MSE-based separation module improve upon the Basic Model, with the MSE-based variant achieving the highest average accuracy among the single-component models (76.93\%). Combining MSE with either GRL or Cen further improves performance, reaching 77.56\% and 76.85\%, respectively. In contrast, directly combining GRL and Cen without the MSE module yields only limited gains (75.53\%), suggesting that effective coordination among different regularization strategies is important.

The Full Model, which integrates all components, achieves the highest average accuracy of 79.24\% across the four days, outperforming all ablated variants. These results indicate that the proposed components are complementary and that their joint optimization is essential for achieving robust cross-receiver generalization.

\subsection{Experiment 4: Hyperparameter Sensitivity Analysis}
\begin{figure}[ht]
  \centering
  \subfigure[GRL weight $\lambda_1$]{
    \includegraphics[width=0.28\columnwidth]{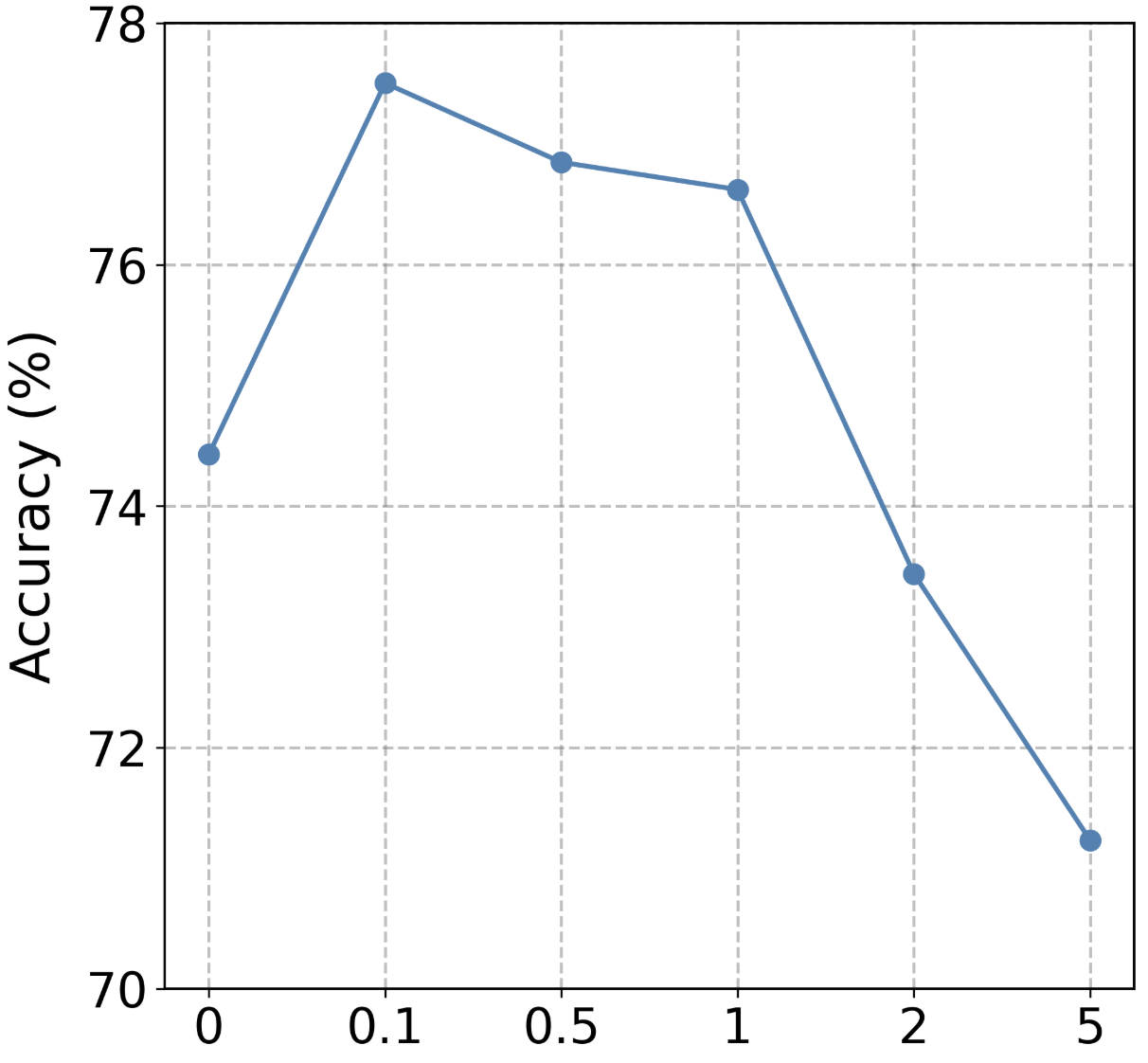}
    \label{fig:alpha_sensitivity}
  }
  \subfigure[Cen weight $\lambda_2$]{
    \includegraphics[width=0.28\columnwidth]{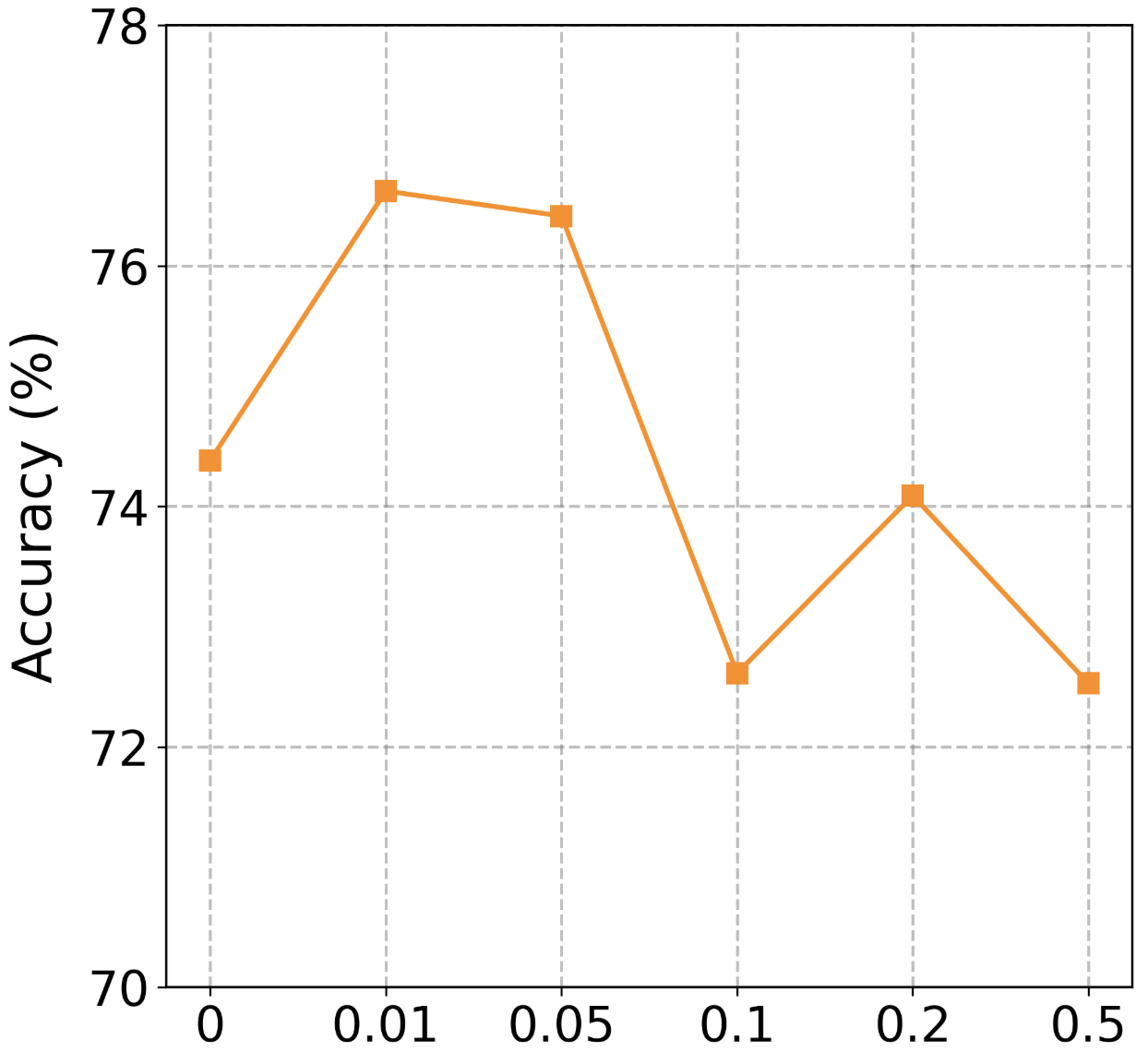}
    \label{fig:beta_sensitivity}
  }
  \subfigure[MSE weight $\lambda_3$]{
    \includegraphics[width=0.28\columnwidth]{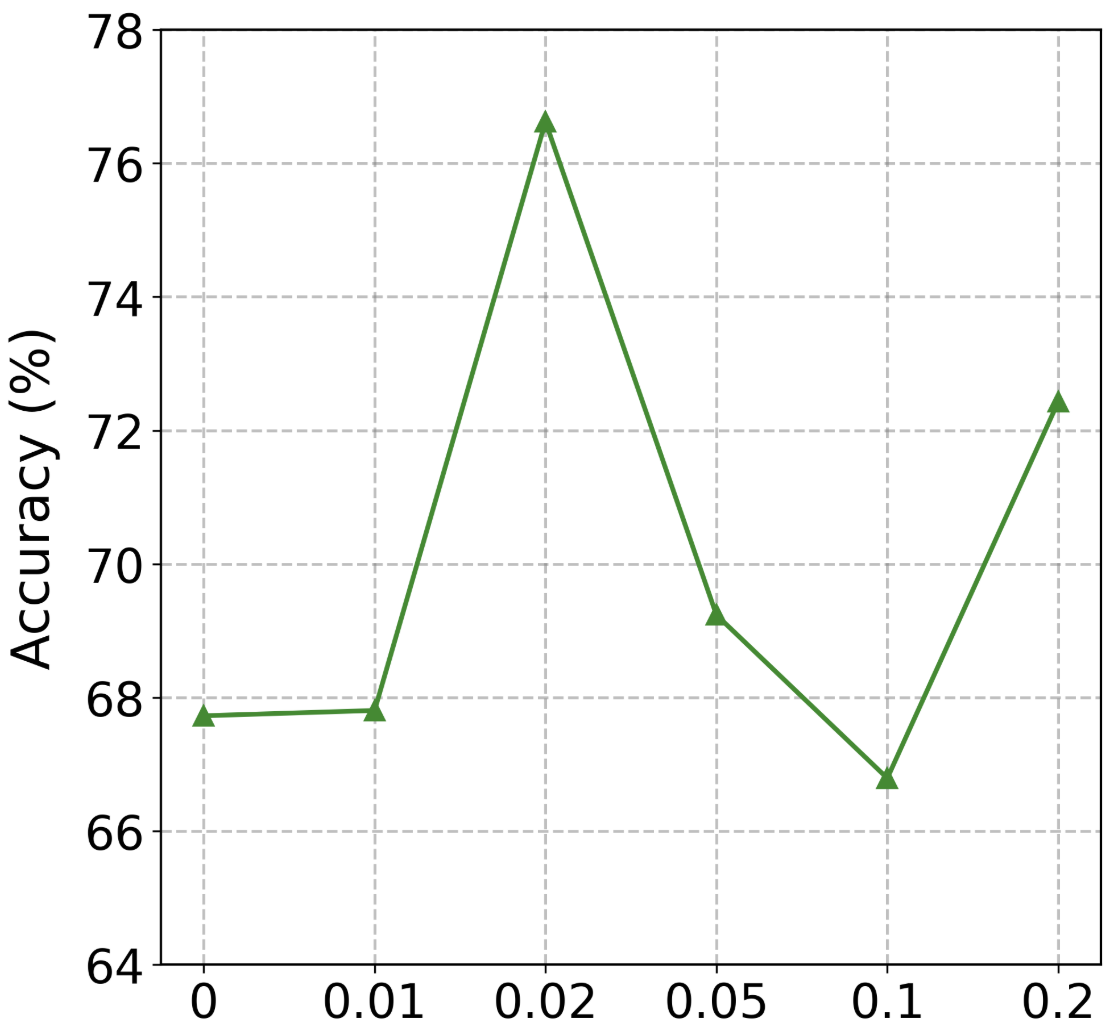}
    \label{fig:gamma_sensitivity}
  }
  \caption{Sensitivity analysis of hyperparameters $\lambda_1$, $\lambda_2$, and $\lambda_3$.}
  \label{fig:hyperparam_sensitivity}
\end{figure}

In this section, we analyze the impact of three key hyperparameters on cross-receiver transmitter identification performance. 
Only one hyperparameter is varied at a time while the others are fixed at their default values, namely $\lambda_1 = 1$, $\lambda_2 = 0.01$, and $\lambda_3 = 0.02$. 
All experiments are conducted under the receiver combination $\{Rx_{(1,1)}, Rx_{(1,19)}, Rx_{(14,7)}, Rx_{(8,8)}\}$.

We first analyze the effect of the GRL weight $\lambda_1$. 
When adversarial learning is disabled ($\lambda_1 = 0$), the average accuracy is 74.43\%. 
Performance improves under moderate adversarial strength and reaches 77.50\% at $\lambda_1 = 0.1$. 
The accuracy remains stable at 76.85\% and 76.62\% for $\lambda_1 = 0.5$ and $\lambda_1 = 1$, respectively. 
However, further increasing $\lambda_1$ to 2 or 5 results in noticeable degradation. 
This indicates that moderate adversarial alignment is beneficial, whereas overly strong domain confusion may suppress discriminative transmitter features.
We then examine the center regularization weight $\lambda_2$. 
Without this constraint, accuracy is 74.38\%. 
A mild regularization strength of 0.01 improves performance to 76.62\%, while stronger regularization (e.g., $\lambda_2 \ge 0.1$) leads to a significant decline. 
These results suggest that appropriate compactness constraints enhance generalization, but excessive regularization restricts model flexibility.
Finally, we investigate the MSE-based separation weight $\lambda_3$. 
Removing the separation loss reduces accuracy sharply to 67.73\%. 
The optimal value is 0.02, where accuracy reaches 76.62\%. 
Both smaller and larger values degrade performance, highlighting the importance of properly balancing feature separation strength.

Overall, the method exhibits moderate sensitivity to $\lambda_1$ and $\lambda_2$, while $\lambda_3$ plays a central role in achieving strong cross-receiver generalization.

\section{Conclusion and Future Work}
\label{sec7}
Most existing RFFI methods rely on the assumption of fixed receivers during both training and testing, overlooking the domain shifts introduced by heterogeneous receiver hardware. This limitation restricts their applicability in practical deployments, where receiver replacement or device migration may be unavoidable.
To address this issue, we propose a cross-receiver generalization framework for RFFI based on feature disentanglement. The proposed model explicitly decomposes input signals into transmitter-specific and receiver-specific representations. A GRL is employed to enforce domain-invariant constraints on transmitter-related features, while center regularization and an MSE-based feature separation loss are introduced to effectively disentangle and constrain receiver-related characteristics.
Comprehensive experiments on a public dataset demonstrate that the proposed method achieves superior identification accuracy and strong generalization ability under both cross-receiver and cross-day evaluation scenarios.
In future work, we plan to extend the framework to more heterogeneous wireless environments with diverse hardware configurations and deployment conditions, further promoting the practical deployment of robust and scalable RFFI systems.

\bibliography{ref}
\bibliographystyle{IEEEtran}

\ifCLASSOPTIONcaptionsoff
  \newpage
\fi

\end{document}